\documentclass[journal]{IEEEtran}
\usepackage[english]{babel}
\usepackage{CJK}
\usepackage{amsmath}
\usepackage{graphicx}
\usepackage{algorithm}
\usepackage{algorithmic}
\usepackage{subfigure}
\usepackage{multirow}
\usepackage{color}
\usepackage{array}
\usepackage{slashbox}
\usepackage{amsmath}
\usepackage{amssymb}
\usepackage{hyperref}
\usepackage[]{caption}
\usepackage{stfloats}
\usepackage{multirow}
\usepackage[table,xcdraw]{xcolor}
\usepackage{array}
\newcommand{\PreserveBackslash}[1]{\let\temp=\\#1\let\\=\temp}
\newcolumntype{C}[1]{>{\PreserveBackslash\centering}p{#1}}
\newcolumntype{R}[1]{>{\PreserveBackslash\raggedleft}p{#1}}
\newcolumntype{L}[1]{>{\PreserveBackslash\raggedright}p{#1}}
\makeatletter
\adddialect\l@ENGLISH\l@english
\makeatother

\begin{document}

\title{Robust 3D Action Recognition through Sampling \\ Local Appearances and Global Distributions}
\author{Mengyuan Liu,~\IEEEmembership{Member,~IEEE,}
        Hong Liu$^{\dag}$,~\IEEEmembership{Member,~IEEE,}
        Chen Chen,~\IEEEmembership{Member,~IEEE}

\thanks{
This work is supported by National Natural Science Foundation of China (NSFC, No.61340046, 61673030, U1613209), Natural Science Foundation of Guangdong Province (No. 2015A030311034), Scientific Research Project of Guangdong Province (NO.2015B010919004), Specialized Research Fund for Strategic and Prospective Industrial Development of Shenzhen City (No. ZLZBCXLJZI20160729020003), Scientific Research Project of Shenzhen City (JCYJ20170306164738129), Shenzhen Key Laboratory for Intelligent Multimedia and Virtual Reality (ZDSYS201703031405467).
}

\thanks{M. Liu is with Faculty of Key Laboratory of Machine Perception, Shenzhen Graduate School, Peking University, Beijing 100871 China, and also with the School of Electrical and Electronic Engineering, Nanyang Technological University, 639798 Singapore (e-mail: liumengyuan@ntu.edu.sg).}
\thanks{H. Liu is with Faculty of Key Laboratory of Machine Perception, Shenzhen Graduate School, Peking University, Beijing 100871 China (e-mail: hongliu@pku.edu.cn). The corresponding author is Hong Liu.}
\thanks{C. Chen is with the Center for Research in Computer Vision at University of Central Florida, Orlando, FL 32816 USA (e-mail: chenchen870713@gmail.com).}}

\markboth{To Appear in IEEE Transactions on Multimedia}%
{Shell \MakeLowercase{\textit{et al.}}: Bare Demo of IEEEtran.cls for Journals}

\maketitle

\begin{abstract}
3D action recognition has broad applications in human-computer interaction and intelligent surveillance. However, recognizing similar actions remains challenging since previous literature fails to capture motion and shape cues effectively from noisy depth data. In this paper, we propose a novel two-layer Bag-of-Visual-Words (BoVW) model, which suppresses the noise disturbances and jointly encodes both motion and shape cues. First, background clutter is removed by a background modeling method that is designed for depth data. Then, motion and shape cues are jointly used to generate robust and distinctive spatial-temporal interest points (STIPs): motion-based STIPs and shape-based STIPs. In the first layer of our model, a multi-scale 3D local steering kernel (M3DLSK) descriptor is proposed to describe local appearances of cuboids around motion-based STIPs. In the second layer, a spatial-temporal vector (STV) descriptor is proposed to describe the spatial-temporal distributions of shape-based STIPs. Using the BoVW model, motion and shape cues are combined to form a fused action representation. Our model performs favorably compared with common STIP detection and description methods. Thorough experiments verify that our model is effective in distinguishing similar actions and robust to background clutter, partial occlusions and pepper noise.
\end{abstract}

\IEEEpeerreviewmaketitle

\section{Introduction}
Determining how to recognize actions (e.g., hugging, hand waving, smoking) accurately in a cost-effective manner remains one of the main challenges for applications such as human-computer interaction (HCI), content-based video analysis and intelligent surveillance \cite{Hu2014Real,Mastorakis2014Fall,xu2017action,hou2017tube}.
Previous methods \cite{dollar2005behavior,laptev2008learning,liu2014learning,liu2014action} have used conventional color cameras to record actions as RGB sequences and developed distinctive action representations to improve the recognition accuracy.
However, action recognition using RGB sequences continues to be challenging due to problems such as different lighting conditions, background clutter and occlusions.

\begin{figure}[t]
\centering
\includegraphics[width=1\linewidth]{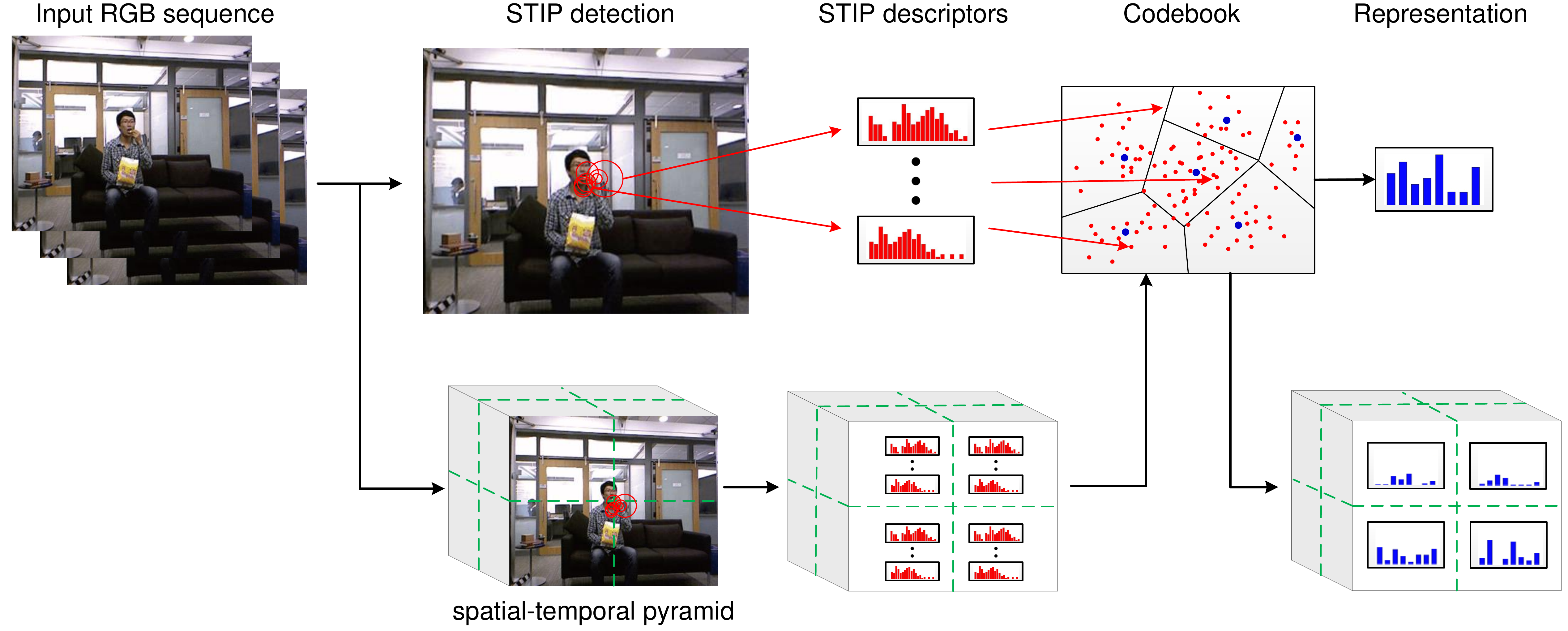}
\captionsetup{font={scriptsize}}
\vspace{-1.5em}
\caption{Flowchart of the traditional STIPs+BoVW+STP framework, where STIPs denotes spatial-temporal interest points, BoVW denotes Bag-of-Visual-Words model and STP denotes spatial-temporal pyramid. The BoVW model refers to the process of aggregating a set of STIPs into a representation. (Best viewed in color)
}\label{1}\vspace{-0em}
\end{figure}

With rapid advances of imaging technology in capturing depth information in real time, there has been a growing interest in solving action recognition problems by using depth data from depth cameras \cite{zhang2017action,chen2017multi,liu2017enhanced}, particularly the cost-effective Microsoft Kinect RGB-D camera. Compared with conventional RGB data, depth data is more robust to changes in lighting conditions because depth values are estimated by infrared radiation without relating it to visible light. Subtracting the foreground from background clutter is easier when using depth data, as the confusing texture and color information from background is ignored. In addition, RGB-D cameras (e.g., Kinect) provide depth maps with appropriate resolution and accuracy, which provide three-dimensional information on the structure of subjects/objects in the scene \cite{ni2013rgbd}.
Using depth data from the Kinect camera, many action recognition systems have been developed \cite{Ren2013Robust,Wang2015Superpixel}.

\begin{figure*}[t]
\centering
\includegraphics[width=1\linewidth]{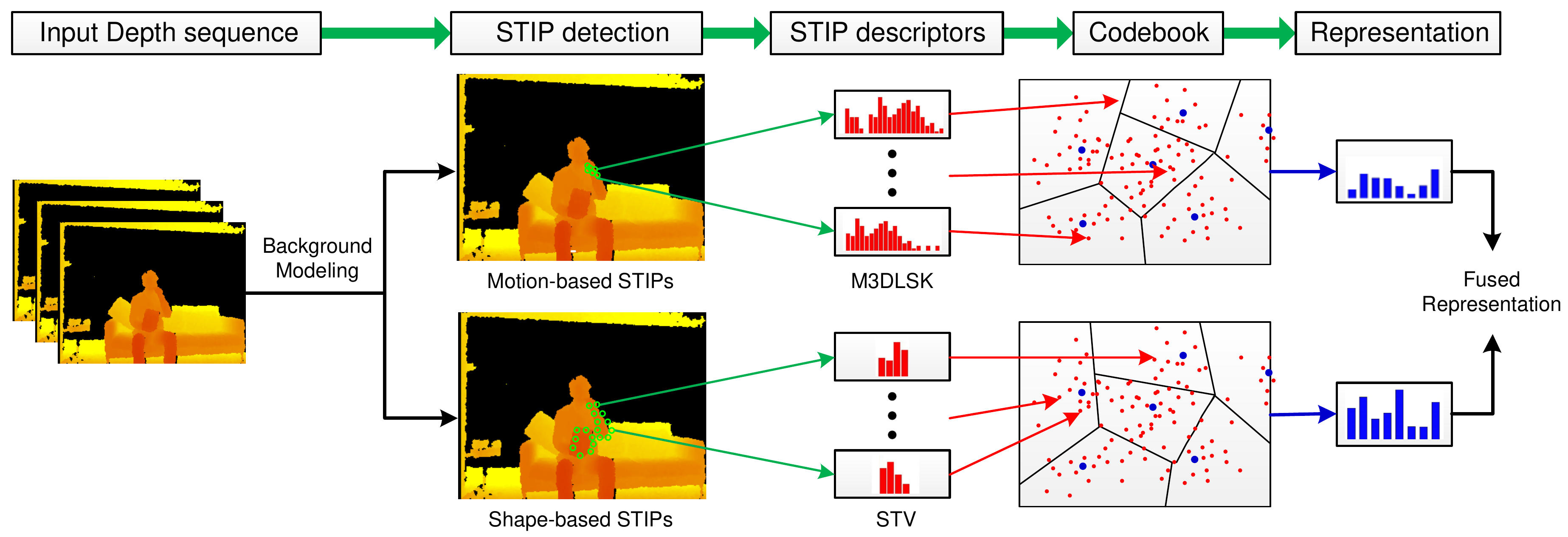}
\captionsetup{font={scriptsize}}
\vspace{-1.5em}
\caption{Two-layer BoVW model for representing 3D actions.
In the first layer, motion-based STIPs are described by the M3DLSK descriptor to represent the motions of actions.
In the second layer, shape-based STIPs are described by the STV descriptor to represent the shapes of actions.
For each layer, we apply BoVW to characterize local features, leading to a bag of M3DLSK representation and a bag of STV representation, which are further combined to form a fused action representation. (Best viewed in color)
}\label{2}\vspace{-.5em}
\end{figure*}

\textcolor[rgb]{0,0,0}{Among the methods of action recognition based on RGB data, ``STIPs+BoVW+STP" is a widely used framework. It is shown in Fig. \ref{1}.
``STIPs" stands for spatial-temporal interest points. ``STIPs" are robust to partial occlusions and able to work with background clutter \cite{dollar2005behavior,laptev2008learning,wang2009evaluation}. ``BoVW" represents the Bag-of-Visual-Words model, which aggregates a set of local descriptors into a vector, i.e., the video representation. ``STP" stands for spatial-temporal pyramid, which encodes the spatial-temporal distribution of STIPs by roughly splitting the spatial-temporal structure into equal cells.
Recent methods \cite{Cheng2012Human, Zhao2012Combing, xia2013spatio, oreifej2013hon4d} applied the ``STIPs+BoVW+STP" framework to the task of action recognition based on depth data.
The main developments focus on designing a more suitable STIP detector and descriptor to accommodate the properties of depth data.}
There are three major shortcomings in this framework.
First, the process of STIP detection and description suffers from noisy depth data.
Second, previous methods on depth action recognition define local points with salient motion as STIPs, while local points with distinctive shape cues are not detected or are ignored.
Third, an STP with a limited number of pyramid levels cannot precisely describe the spatial-temporal distribution of STIPs, while an STP with many pyramid levels may lead to a high-dimensional action representation, which would result in lower discriminative power and higher time cost for classification.

To solve these problems, this paper presents a two-layer BoVW model (shown in Fig. \ref{2}) for 3D action representation.
This model treats a 3D action as a set of local STIPs.
Local appearances and global distributions of these STIPs are jointly used for action representation.
Specifically, in the first layer, motion-based STIPs are sampled from depth data and then described by M3DLSK descriptor to represent local appearances.
In the second layer, shape-based STIPs are sampled from depth data and described by STV descriptor to represent global distributions.
Our main contributions are three-fold:

1. A new background modeling method is developed to efficiently remove the background from depth data. As compared to other background subtraction approaches, our method is more effective and efficient.

2. Motion-based and shape-based STIPs are sampled from depth data to explicitly encode the motion and shape (appearance) information to leverage the complementary nature of these two action cues. Particularly, shape-based STIPs provide complementary information for motion-based STIPs in two cases: a) human-object interactions with similar actions and different objects, b) different types of actions with small motions. In both cases, shape-based STIPs can describe shapes of objects or action performers, in order to provide distinctive cues for action recognition.

3. 3DLSK has been used for motion description of depth videos for the first time. We go beyond the 3DLSK descriptor by extending it to multi-scale 3DLSK (M3DLSK) for capturing more detailed local motion information. To capture spatial-temporal relationships (global structure information) among STIPs, a spatial-temporal vector (STV) is proposed to encode the global distribution of motion-based STIPs as well as shape-based STIPs. By fusing the M3DLSK and STV descriptors, the fused feature presentation is endowed with the capacity of capturing both local and global motion and shape cues.

The rest of this paper is organized as follows. First, related methods on 3D action recognition are reviewed in Section II.
Second, the foreground extraction and STIP detection methods are presented in Section III.
Third, M3DLSK and STV descriptors are described in Section IV. Fourth, experiments and analysis are reported in Section V.
Finally, conclusions and future work are summarized in Section VI.

\vspace{-1em}
\textcolor[rgb]{0,0,0}{
\section{Related Work}
According to the type of input data, 3D action recognition methods can be categorized into skeleton-based approaches \cite{Wang2014Mining,Chen2016A,Amor2016Action,Anirudh2016Elastic}, hybrid approaches \cite{Sung2012Unstructured,Koppula2013Learning,Liu20163D,Zhang2016RGB,Jalal2017Robust} and depth-based approaches \cite{Cheng2012Human,Zhao2012Combing,yang2012recognizing,xia2013spatio,oreifej2013hon4d,li2010action,Slama2014Grassmannian,wang2015convnets,Cho2015Volumetric,Zhang2016Local,Wang2015Action,veeriah2015differential,Shahroudy2016NTU,Yang2016Super}.
The applications of skeleton-based and hybrid approaches are limited, since skeleton data may be inaccurate when a person is not directly facing the camera. Moreover, skeleton data cannot be obtained in applications such as hand gesture recognition.
According to the type of feature, depth-based approaches can be divided into image-based approaches \cite{wang2015convnets,Wang2015Action,veeriah2015differential,Shahroudy2016NTU}, silhouette-based approaches \cite{yang2012recognizing,li2010action,Cho2015Volumetric,Slama2014Grassmannian} and STIP-based approaches \cite{Cheng2012Human,Zhao2012Combing,xia2013spatio,oreifej2013hon4d,Zhang2016Local,Yang2016Super}. Image-based approaches directly use the original image as a feature and train convolutional neural networks to obtain an end-to-end system \cite{wang2015convnets,Wang2015Action}.
Recently, differential Recurrent Neural Networks (RNN) \cite{veeriah2015differential} and part-aware Long Short-Term Memory (LSTM) \cite{Shahroudy2016NTU} have been proposed to model temporal relationships among frames. Although high accuracies have been achieved, these methods need large labeled datasets and incur high time cost for training.
Silhouette-based approaches extract features from silhouettes, and then some methods model the dynamics of the action explicitly using statistical models \cite{li2010action,Cho2015Volumetric,Slama2014Grassmannian}. However, these methods have difficulty handling partial occlusions and background clutter.}

\textcolor[rgb]{0,0,0}{Belonging to the STIP-based approaches, our work contains three stages: detecting STIPs, describing STIPs and encoding global relationships.
To detect STIPs, Laptev designed a detector that defines STIPs as local structures in which the illumination values show large variations in both space and time \cite{laptev2005space}.
Four of the subsequently developed STIP detectors, namely, the Harris3D detector, Cuboid detector, Hessian detector and dense sampling, were evaluated by Wang \textit{et al.} \cite{wang2009evaluation}.
For describing STIPs, Cuboid \cite{dollar2005behavior}, HOG/HOF \cite{laptev2008learning} and HOG3D \cite{klaser2008spatio} descriptors were also evaluated in \cite{wang2009evaluation}.
The STIP detectors and descriptors can be directly applied to 3D action recognition \cite{Cheng2012Human,Zhao2012Combing}.
However, these methods generally do not work well due to noise, e.g., space-time discontinuities, in depth data \cite{ni2013rgbd}.
Recently, some STIP detectors and descriptors have been specially designed for depth data \cite{xia2013spatio,oreifej2013hon4d,Zhang2016Local,Yang2016Super}.
However, they still suffer from the effects of depth noise. Moreover, the salient shape cues are usually ignored by these methods.
For encoding global relationships among STIPs, the spatial-temporal pyramid (STP) \cite{laptev2008learning} is widely used.
Each video is equally divided into cells, and representations of all cells are concatenated to form the final video representation.
However, STP cannot achieve both high spatial-temporal resolution and distinctive action representation with low dimensions.
To solve the above problems, a robust and descriptive method for representing 3D actions from noisy depth data is necessary.}

\section{STIP Detection}
\textcolor[rgb]{0,0,0}{A foreground extraction method is first proposed to remove background clutter; thereafter, using this method, noise from the background is removed. Then, motion and shape information are jointly used to eliminate noise disturbances from the foreground. Motion-based STIPs and shape-based STIPs are detected to capture salient motion and shape cues, which are useful for distinguishing similar actions.}

\subsection{Foreground Extraction}
\begin{figure}[t]
\begin{center}
\includegraphics[width=1\linewidth]{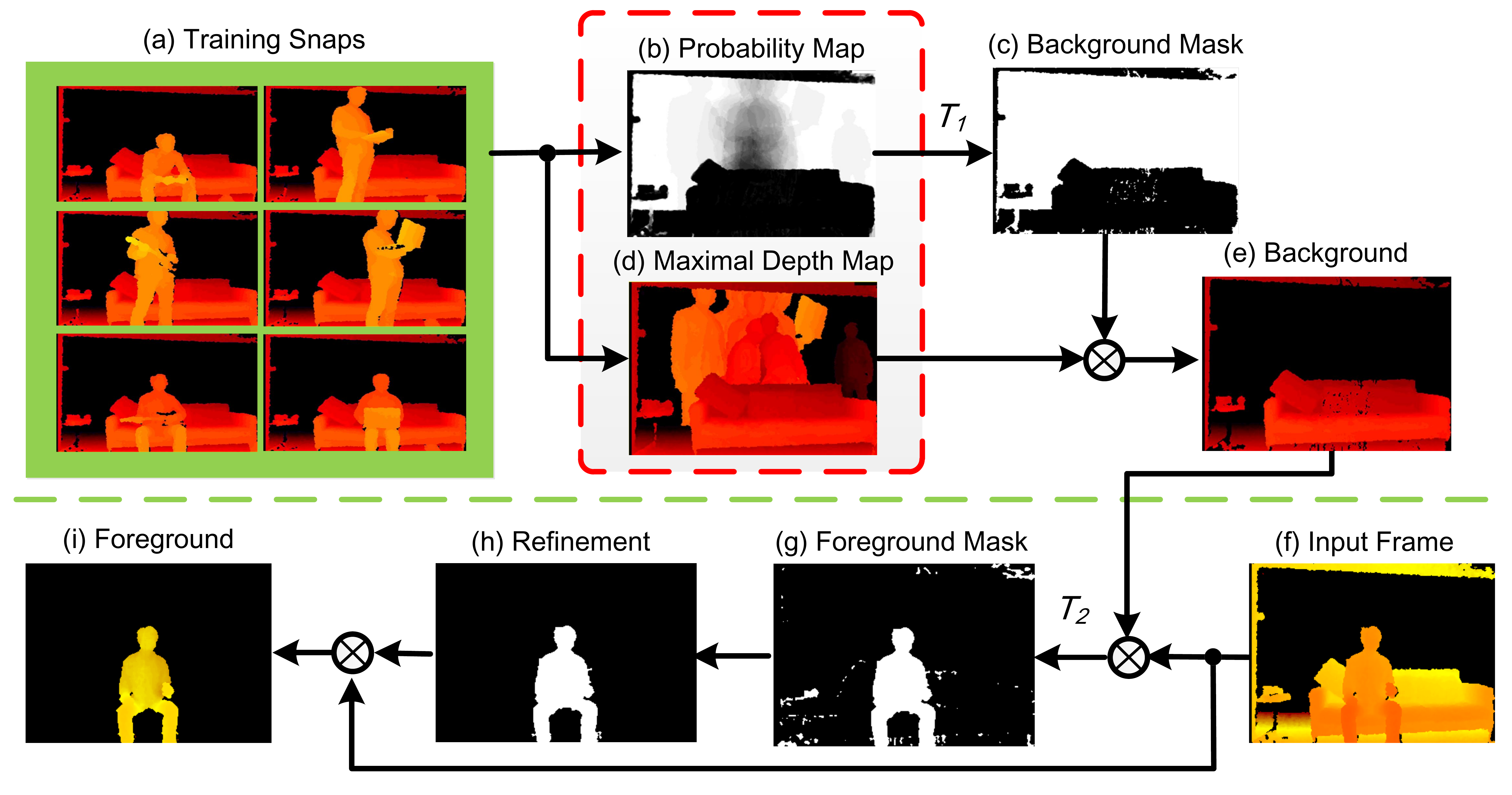}
\end{center}
\vspace{-1em}
\captionsetup{font={scriptsize}}
\caption{Background modeling and foreground extraction. The background model is built by updating a maximal depth map and a probability map. The maximal depth map records the observed maximal depth value of each pixel. The probability map indicates the probability of each pixel belonging to the far-field background.}\label{3}\vspace{-0em}
\end{figure}

Our proposed depth foreground extraction method includes two parts: background modeling and foreground extraction. The effective foreground includes a human body (assuming one person in the scene) and objects that she/he interacts with.
In Fig. \ref{3} (a), the person and objects (e.g., computer) belong to the foreground.
In Fig. \ref{3} (e), the background is built by observing several training frames, each of which can be regarded as the background occluded by the foreground.
The purpose of background modeling is to recover the background regions that are partially occluded by the foreground.

According to the distance from the camera, the background can be classified into far-field background and near-field background.
If the distance of the far-field background is beyond the sensing range of the camera, the depth value of the far-field background cannot be obtained.
In the depth map provided by the Kinect device \cite{Zhang2012Microsoft}, the depth value of the far-field background is set to zero (e.g., large black areas in the training examples shown in Fig. \ref{3} (a)).
Thereafter, we propose the concept of a probability map, which is used to describe the probability of each pixel position belonging to the far-field background.
For each position, we count the number of times that the depth value is equal to zero. Then, the percentage of zero values is used as the probability.
Concretely, given a set of depth frames $\{I^n\}_{n=1}^N$ with $N$ frames, we define a probability map $P$ as:
\begin{equation}\label{5}
\setlength\abovedisplayskip{2pt}
P = \Big\{ P(x,y) | P(x,y) =  \frac{1}{N} \sum_{n=1}^N{ \delta \big(I^n(x,y)\big) } \Big\},
\setlength\belowdisplayskip{2pt}
\end{equation}
where $I^n$ is the $n$-$th$ frame, $(x,y)$ denotes the location of one pixel and the function $\delta$ is defined as:
\begin{equation}\label{6}
\setlength\abovedisplayskip{2pt}
\begin{aligned}
\delta(a)= \left\{
\begin{array}{rl}
1,& if \:\: a=0\\
0,& otherwise
\end{array}
\right.
\end{aligned}.
\setlength\belowdisplayskip{2pt}
\end{equation}
If the probability is large, the location most likely belongs to the far-field background.
Note that the non-zero depth values indicate occlusions, i.e., foreground objects.

\begin{figure*}[t]
\begin{center}
\includegraphics[width=1\linewidth]{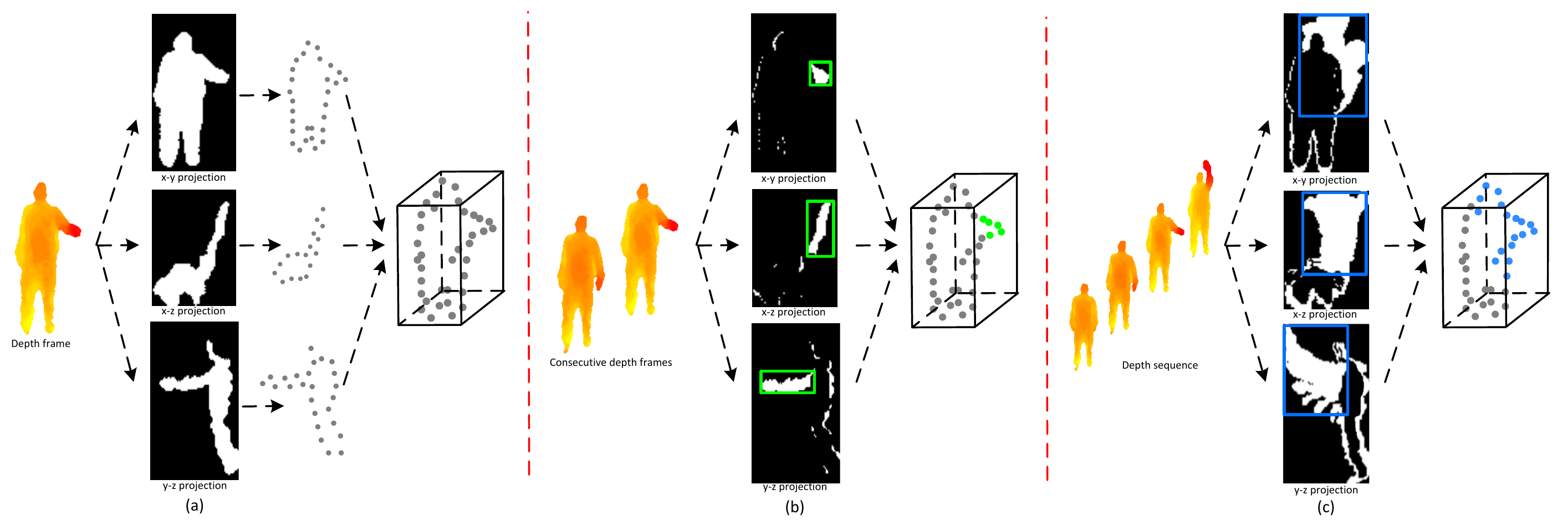}
\end{center}
\vspace{-1em}
\captionsetup{font={scriptsize}}
\caption{STIP detection Methods. (a) STIP candidates are generated by sampling points on the contours of projected maps. (b) Motion-based STIPs are selected from the candidates using the bounding boxes of the inter-frame motion. (c) Shape-based STIPs are selected from the candidates using the bounding boxes of the accumulated inter-frame motion.}\vspace{-1.5em}\label{11}
\end{figure*}

Since the near-field background is within the sensing range of the camera, the depth value of the near-field background is non-zero. For example, in Fig. \ref{3} (a), the sofa in the depth images is considered part of the near-field background.
At the same time, the distance between the camera and the near-field background is longer than the distance between the camera and the foreground; that is to say, the depth value in any position of the near-field background is larger than those of the foreground objects. Based on this observation, we propose a novel maximum depth map, which is used to record the maximum depth value of each position in a depth sequence. Specifically, given a set of depth frames $\{I^n\}_{n=1}^N$, we define the maximal depth map $M$ as:
\begin{equation}\label{4}
\setlength\abovedisplayskip{2pt}
  M = \Big\{ M(x,y) | M(x,y) = max\big\{ I^n(x,y)\big\}_{n=1}^N \Big\}.
\setlength\belowdisplayskip{2pt}
\end{equation}
In the region of the near-field background, the maximum depth value can describe the near-field background.

Combining the far-field and near-field background, the depth value at location $(x,y)$ of the background model $B$ is defined as:
\begin{equation}\label{6}
\setlength\abovedisplayskip{2pt}
\begin{aligned}
B(x,y)= \left\{
\begin{array}{rl}
0,& if \:\: P(x,y)>T_1\\
M(x,y),& otherwise
\end{array}
\right.
\end{aligned}.
\setlength\belowdisplayskip{2pt}
\end{equation}
Applying the background subtraction algorithm, we can obtain the foreground region, which is further binarized by a threshold $T_2$. Since the presence of one person is assumed in each depth frame, we use the two-pass algorithm \cite{Stockman2001Computer} to extract the connected region with the largest area in the refined foreground region. In this way, we remove the connected regions with small areas, which are produced by the noise.
The background modeling method proposed in this paper is different from traditional background subtraction methods, e.g., Mean filter \cite{Benezeth2008Review} and ViBe \cite{Barnich2011ViBe}, which distinguish moving objects (called the foreground) from static (or slow-moving) parts of the scene (called the background). These methods tend to merge the foreground into the background when the foreground is static for a period of time. In other words, these methods focus on extracting the moving parts. In contrast, our method can extract the whole human body and the objects that the person interacts with.
Based on the foreground, we extract STIPs from both motion regions and static regions in the next section.

\subsection{Motion-based and Shape-based STIPs}
We combine 3D shape and 3D motion cues to detect STIPs, where the shape cue is used to obtain the STIP candidates, and the motion cue is used to select the target STIPs from the candidates. We obtain two types of target STIPs: motion-based STIPs for describing motion information and shape-based STIPs for describing shape information.

The first step is to obtain the STIP candidates. Studies on 2D silhouette-based action recognition have verified that shape information of human bodies can be reflected by contours. To make full use of depth information, Li et al. \cite{li2010action} projected each depth frame onto three orthogonal Cartesian planes and sampled a number of points at equal distance along the contours of the projections.
The distance along the contours is denoted by $\lambda$, which determines the density of sampled points.
As shown in Fig. \ref{11} (a), 2D points on three projected maps are combined with corresponding depth values to form 3D points, which are also called STIP candidates (colored in gray).
Suppose we obtain a set of STIP candidates $\mathcal{S}$, where $\mathbf{v}=(x,y,z,f)$ stands for a single STIP, $x$ and $y$ represent the horizontal and vertical coordinates, $z$ is the depth value and $f$ is the frame number.
Note that the projections onto the \textit{xz} and \textit{zy} planes are interpolated to solve the problem of strip-like noise, which results from discontinuous depth values.

In the second step, we use inter-frame motion to select motion-based STIPs from the STIP candidates.
Let ${F}^f$ denote the foreground of the $f$-$th$ frame. We project it onto three maps: $map_{xy}^f, map_{xz}^f$ and $map_{yz}^f$. For the $\phi \in \{xy,xz,yz\}$ view, we calculate the inter-frame motion region by:
\begin{equation}\label{13}
\setlength\abovedisplayskip{2pt}
R_{\phi}^f = |map_{\phi}^f - map_{\phi}^{f-1}|>\epsilon,
\setlength\belowdisplayskip{2pt}
\end{equation}
where $R_{\phi}^f$ is a binary map and $\epsilon$ is a threshold. Similar to \cite{yang2012recognizing}, we set $\epsilon$ to $50$.
As a preprocessing step, we apply the ``closing" morphological operator \cite{SerraImage} on $R_{\phi}^f$ to generate a set of connected regions.
To reduce the effect of noise, we use the two-pass algorithm \cite{Stockman2001Computer} to label the connected regions. Each connected region whose area is larger than $0.8$ times the largest area is reserved as a refined motion region.
As shown in Fig. \ref{11} (b), a bounding box is formed on each orthogonal Cartesian plane to indicate the motion region.
We selected a subset of the candidates as motion-based STIPs (colored in green), whose projected locations are inside the three bounding boxes.

\begin{figure}[t]
\begin{center}
\includegraphics[width=1\linewidth]{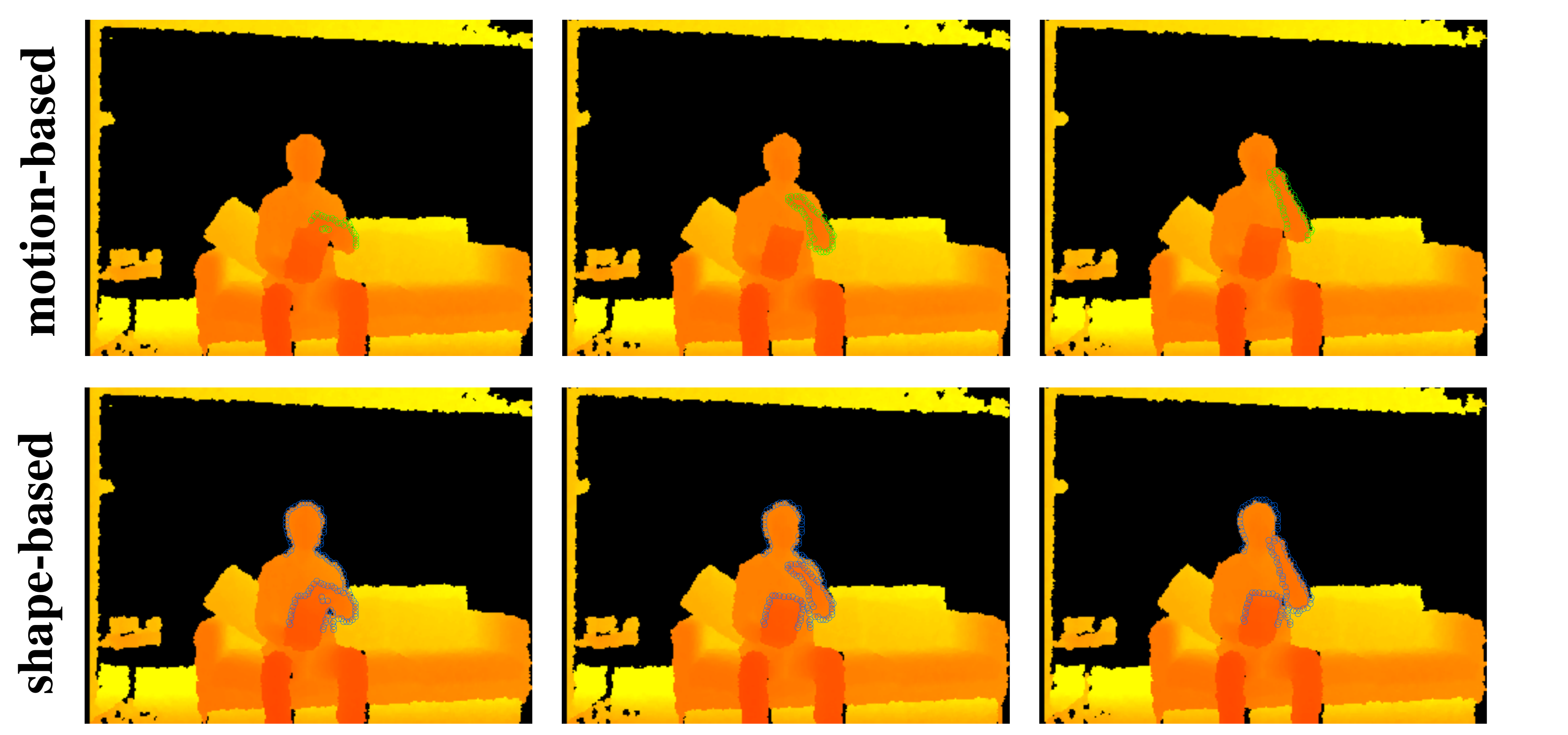}
\end{center}
\vspace{-1.0em}
\captionsetup{font={scriptsize}}
\caption{Examples of motion-based and shape-based STIPs}\label{12}
\end{figure}

In the third step, we use accumulated inter-frame motion information over a depth action sequence to select shape-based STIPs from the STIP candidates.
We observe that the STIP candidates that fall out of the range of motion are weakly related to the action.
Take the action of ``waving a hand'' in Fig. \ref{11} as an example. The range of motion is mainly the entire range of area that the hand reaches. The STIP candidates that are located in the lower part of the human body are not directly related to the action. In other words, the states of these STIPs (e.g., standing or sitting when a person performs the action ``waving a hand'') do not affect the type of action.
Therefore, we use the accumulated inter-frame motion to represent the approximate range of motion and then select shape-based STIPs.
Let $A_{\phi}$ denote the accumulated inter-frame motion map on the $\phi$ view:
\begin{equation}\label{15}
\setlength\abovedisplayskip{2pt}
A_{\phi}=\sum\nolimits_{f}{R_{\phi}^f},
\setlength\belowdisplayskip{2pt}
\end{equation}
Following a similar procedure to that in the second step, we select a subset of the candidates as shape-based STIPs (colored in blue), which are shown in Fig. \ref{11} (c).

Fig. \ref{12} shows the sampled motion-based and shape-based STIPs from an action of ``eating snacks" in three frames.
To facilitate observation, the parameter $\lambda$ (i.e., the distance between two sampling points along the contours \cite{li2010action}) is set to three pixels.
As seen in Fig. \ref{12}, the motion-based STIPs are mainly located in the motion region of the foreground, reflecting the salient motion information.
The shape-based STIPs are mainly located in the overall range of foreground motion, thus showing strong relations to the action. It is also evident that the shape-based STIPs can represent the shapes of objects (e.g., the snack bag in Fig. \ref{12}), which are essential for distinguishing actions that involve interactions with different objects using similar motions.

\section{STIP Description}
\subsection{Scale-adaptive M3DLSK Descriptor}
Originally designed for color data, 3DLSK \cite{seo2011action} can implicitly capture the local structure of one STIP by estimating a local regression kernel using nearby points. Here, we introduce the original 3DLSK to describe local structures in depth data.
Different from the color data, the pixel value in depth data reflects its distance to the camera.
Using this characteristic, we propose a scale-adaptive 3DLSK, which suffers less from the effect of scale changes.
Since the depth data lack textural information, we propose a multi-scale 3DLSK (M3DLSK), which captures richer local information than 3DLSK.

Suppose an STIP $\mathbf{v}=(x,y,z,f)$ belongs to the motion-based STIPs. Let $\Omega_r$ denote a spatial-temporal cuboid around the STIP, where the parameter $r$ determines the scope of the cuboid. Following the default parameter settings in \cite{seo2011action}, we use the 3DLSK descriptor to describe the cuboids of motion-based STIPs. Then, we apply the Bag-of-Visual-Words (BoVW) model, which clusters the descriptors into centers and returns the frequency histogram of the clustered descriptors. All descriptors are characterized as a histogram $H_{r,k_1}$ where $k_1$ is the number of clusters.

To achieve scale invariance of the spatial-temporal cuboid, we associate the parameter $r$ with the depth value of the STIP.
Since the depth value of a single STIP is not stable, we use an average depth value as the indicator of the real depth value.
Specifically, we first select a cuboid $\Omega$. Then, we calculate the average depth value $\bar{z}$ as:
\begin{equation}\label{20}
\setlength\abovedisplayskip{2pt}
\bar{z}= {\sum\nolimits_{(x,y,f)\in \Omega_{3}}{z}} \Big/ {\sum\nolimits_{(x,y,f)\in \Omega_{3}}{num\big(z\big)}},
\setlength\belowdisplayskip{2pt}
\end{equation}
where $(x,y,f)$ is a point; the function $num(z)$ returns one when $z>0$, and otherwise returns zero; and $\bar{z}$ records the average depth value of the foreground.
Note that we directly discard the cuboid whose depth values are all equal to zero.
The adaptive scale is formed as:
\begin{equation}\label{21}
\setlength\abovedisplayskip{2pt}
\hat{r}=\frac{\bar{z}_0}{\bar{z}} \cdot r,
\setlength\belowdisplayskip{2pt}
\end{equation}
where $\bar{z}_0$ is the average depth value of the foreground from the training set.
Applying 3DLSK to cuboids of different sizes will generate descriptors of different lengths, which cannot serve as inputs to the BoVW model.
Therefore, we normalize the cuboid with a scale of $\hat{r}$ to a normalized scale of $r$.
In the following, we use the parameter $r$ to represent the scale and adopt the adaptive scale adjustment method.

Despite of the use of an adaptive scale, we argue that a cuboid in a single scale cannot capture detailed information about local structures.
Therefore, we propose a bag of multi-scale 3DLSK (M3DLSK) representation as $H_{motion}=\big[H_{r^l,k_1}\big]_{l=1}^L$,
where $H_{motion}$ is a representative-level fusion of 3DLSK with the parameter $r^l$, and $l$ ranges from $1$ to $L$.
Here, $L$ means the number of scales.

\begin{figure}[t]
\begin{center}
\includegraphics[width=1\linewidth]{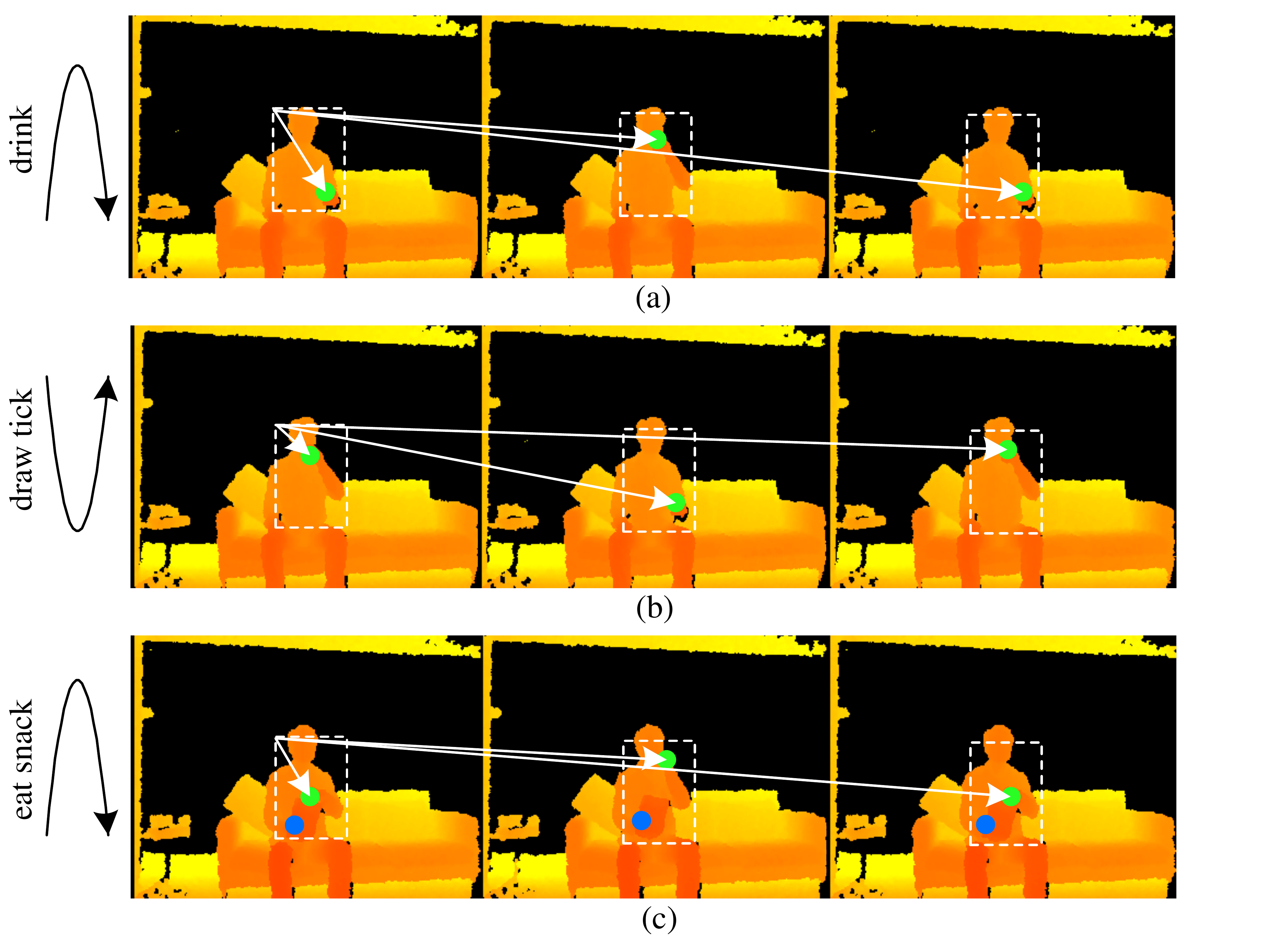}
\end{center}
\vspace{-1.0em}
\captionsetup{font={scriptsize}}
\caption{\textcolor[rgb]{0,0,0}{
Illustration of the complementarity property between STV and M3DLSK.
Green dots stand for motion-based STIPs and blue dots stand for shape-based STIPs.
White rectangles indicate the motion scope.
The origin of a sequence is defined as the upper-left corner of the rectangle in the first frame.
We link the origin to each STIP to form an STV descriptor, which is denoted by a white vector.
Three actions are shown in (a), (b), (c), whose local structures along the motion trajectories are similar.
Therefore, these actions cannot be distinguished by the motion-based STIPs described by the M3DLSK descriptor.
To distinguish action (a) and action (b), STV descriptors are extracted to capture their distinct spatial-temporal distributions of motion-based STIPs.
To distinguish action (a) and action (c), STV descriptors are extracted from shape-based STIPs to capture the shape of the object (i.e., the snack bag). (Best viewed in color)}
}\label{27}\vspace{.8em}
\end{figure}

\subsection{STV Descriptor}
\vspace{-.5em}
\textcolor[rgb]{0,0,0}{
The proposed M3DLSK descriptor can efficiently capture local structures of motion-based STIPs.
However, this method ignores the spatial-temporal relationships among STIPs, which results in ambiguities when distinguishing actions with similar numerical distributions of STIPs.
In Fig. \ref{27} (a) and (b), two actions are shown that contain similar motion-based STIPs (green dots).
These two actions can barely be distinguished by using M3DLSK to describe the local structures of motion-based STIPs.
Since the main difference between actions (a) and (b) lies in the spatial-temporal distribution of STIPs, we extract a spatial-temporal vector (STV) descriptor for each STIP and use these STV descriptors to distinguish between the two actions.
Note that the STV descriptors are denoted by white vectors, which are different for action (a) and action (b).}

\textcolor[rgb]{0,0,0}{
In addition to encoding the global distribution of motion-based STIPs, STV descriptors can be used to describe shape-based STIPs. Especially for human-object interaction, shape-based STIPs play an important role in describing shapes of objects.
In Fig. \ref{27} (a) and (c), two actions that contain similar motion-based STIPs are shown and the main difference between them lies in the objects (i.e., the snack bag). Thereafter, we detect shape-based STIPs (blue dots) and describe them using STV descriptors.
Generally, three actions can be distinguished by combining the M3DLSK descriptor and STV descriptor.
Note that motion-based STIPs can be regarded as a subset of shape-based STIPs, as shown in Fig. \ref{11}. In other words, the spatial-temporal distributions of shape-based STIPs not only contain shape cues but also reflect the global relationships among motion-based STIPs. Thereafter, STV descriptors are created based on shape-based STIPs.}

\textcolor[rgb]{0,0,0}{
Suppose $\hat{S}$ denotes a set of shape-based STIPs, and $\mathbf{v}=(x,y,z,f)$ denotes one STIP from $\hat{S}$.
An origin is defined as:
\begin{equation}\label{24}
\setlength\abovedisplayskip{2pt}
\mathbf{o}=(\mathop{min}\limits_{\forall \mathbf{v}\in \hat{S}}\{x\}, \mathop{min}\limits_{\forall \mathbf{v}\in \hat{S}}\{y\}, \mathop{min}\limits_{\forall \mathbf{v}\in \hat{S}}\{z\}, \mathop{min}\limits_{\forall \mathbf{v}\in \hat{S}}\{f\}),
\setlength\belowdisplayskip{2pt}
\end{equation}
where $\mathop{min}\limits_{\forall \mathbf{v}\in \hat{S}}\{x\}$ returns the minimum x-axis value of all STIPs from $\hat{S}$. For STIP $\mathbf{v}$, the corresponding STV descriptor is defined as a 4D vector that points from the origin $\mathbf{o}$ to STIP $\mathbf{v}$.
For an action sequence, we further normalize each dimension of the STV descriptor to $[0,1]$. This step reduces the differences among various performers, e.g., different human body sizes, which results in different spatial-temporal distributions of STIPs. Then, STV descriptors are used as inputs to the BoVW model to form the shape-based representation $H_{shape}=H_{k_2}$ where $k_2$ is the number of clusters.}

\textcolor[rgb]{0,0,0}{We concatenate the feature representations generated by the M3DLSK and STV descriptors to form a fused representation $H = [ H_{motion}, H_{shape} ]$, where both motion and shape information are encoded.}

\begin{figure}[t]
\begin{center}
\includegraphics[width=.95\linewidth]{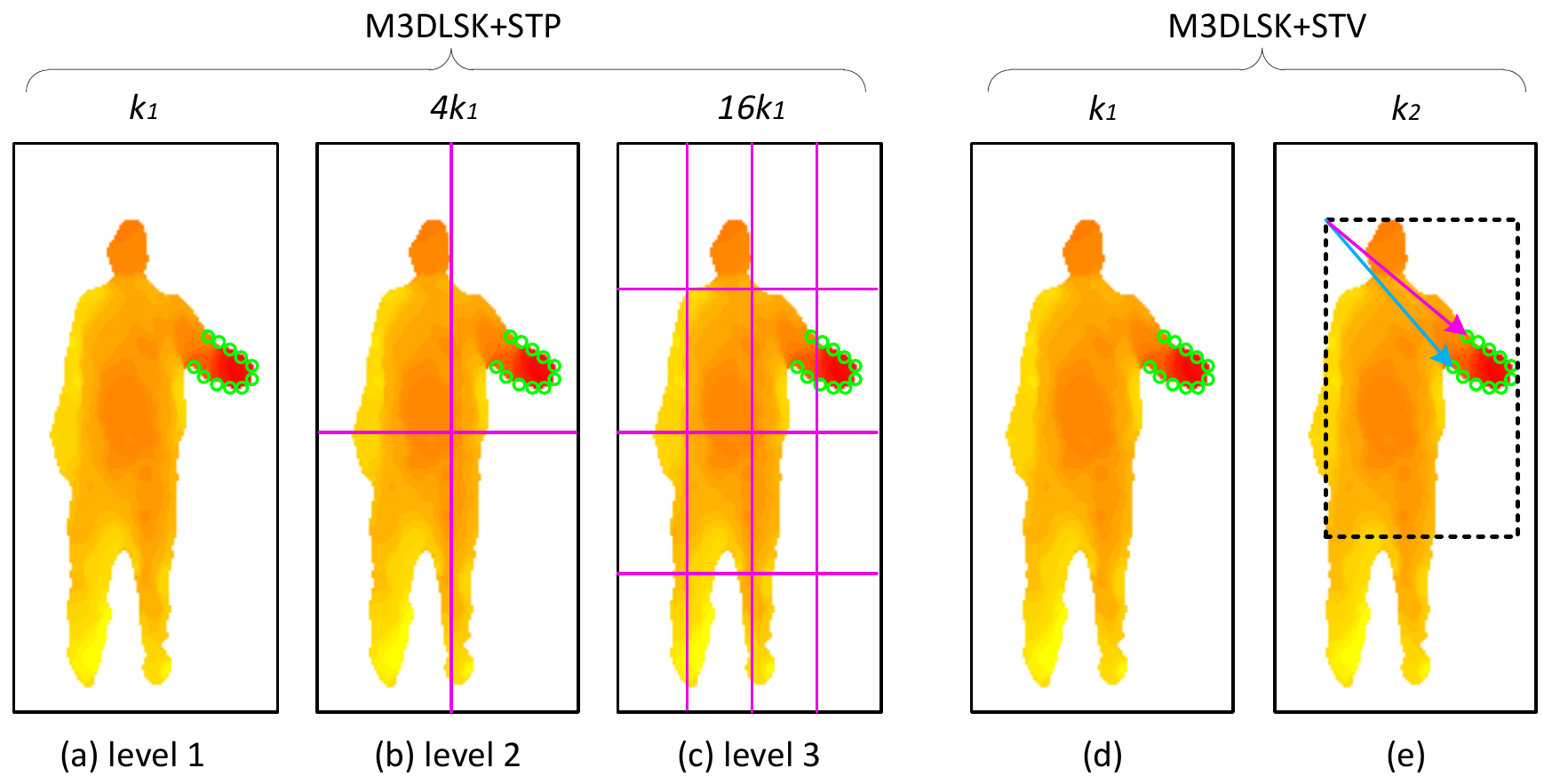}
\end{center}
\vspace{-1.0em}
\captionsetup{font={scriptsize}}
\caption{\textcolor[rgb]{0,0,0}{Comparison between STP and STV}}\label{1213}
\end{figure}

\textcolor[rgb]{0,0,0}{An STP with three levels is illustrated in Fig. \ref{1213}, where the temporal and spatial axes are divided into $1\times1\times1$ cells (Fig. \ref{1213} (a)), $1\times2\times2$ cells (Fig. \ref{1213} (b)) and $1\times4\times4$ cells (Fig. \ref{1213} (c)). The STIPs in each cell are described by the M3DLSK descriptor and then represented by a $k_1$-dimensional vector using the BoVW model.
The final representation using M3DLSK+STP is formulated by concatenating vectors from all cells.
The representation that is obtained using M3DLSK+STP is a $21k_1$-dimensional vector.
The problem of STP is that the spatial-temporal distributions of STIPs in each cell are not encoded.
A possible solution is to divide the volume into smaller-sized cells.
However, this method will lead to a high-dimensional vector, which will result in higher time cost for classification and lower discriminative power.
The merit of the proposed STV is that it can achieve high spatial-temporal resolution resolution while generating a final low-dimensional representation. Taking the STV in Fig. \ref{1213} as an example, the dimension of the final representation is $k_1+k_2$, where the parameter $k_2$ is the number of clusters of spatial-temporal vectors. In Fig. \ref{1213} (e), the spatial-temporal information of nearby STIPs can be distinguished by quantized STVs (vectors colored in blue and red).}

\begin{figure}[t]
\centering
\includegraphics[width=.92\linewidth]{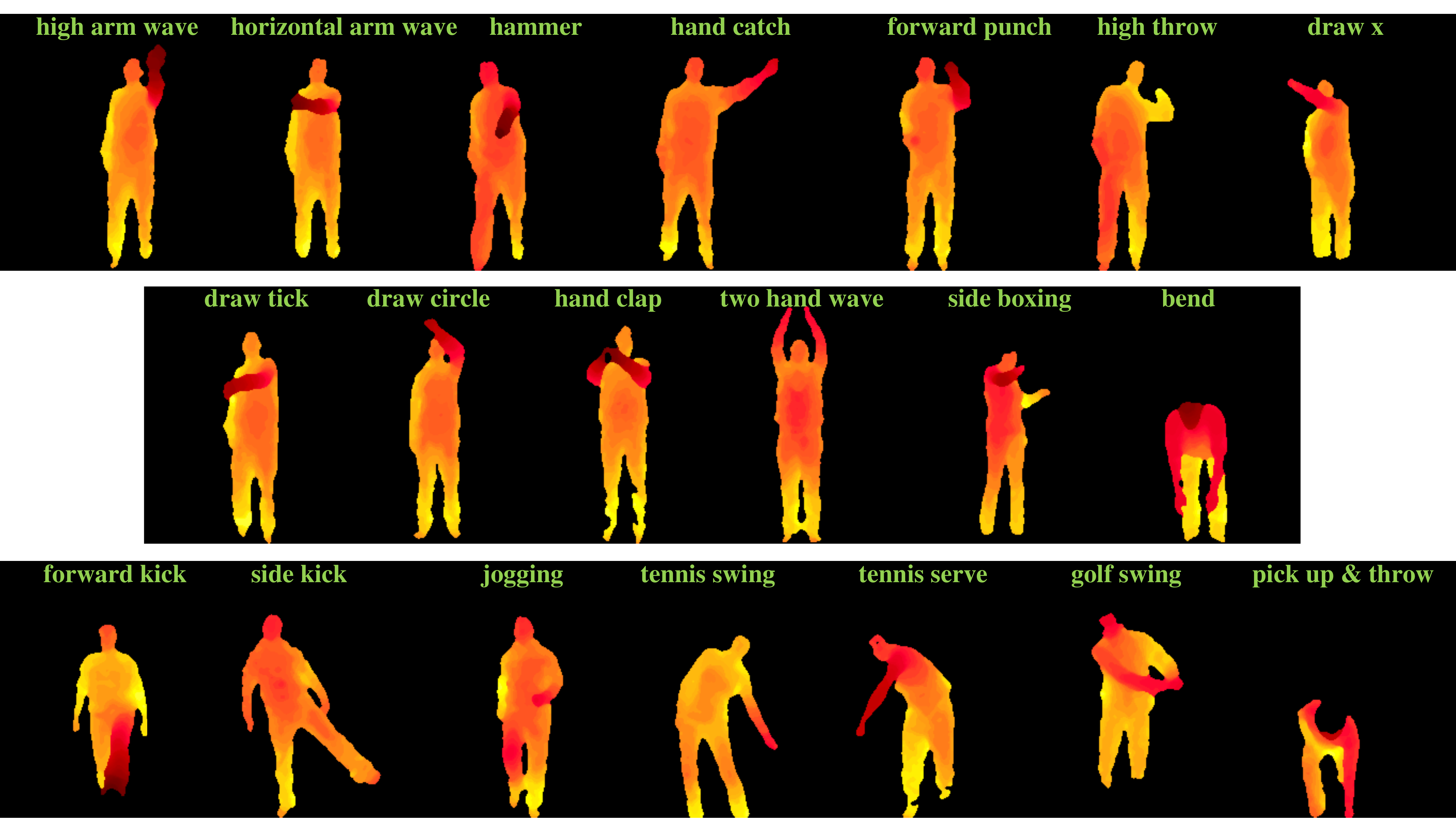}
\captionsetup{font={scriptsize}}
\vspace{-.5em}
\caption{Keyframes from the MSRAction3D dataset
}\label{28}\vspace{-.5em}
\end{figure}

\begin{figure}[t]
\centering
\includegraphics[width=.92\linewidth]{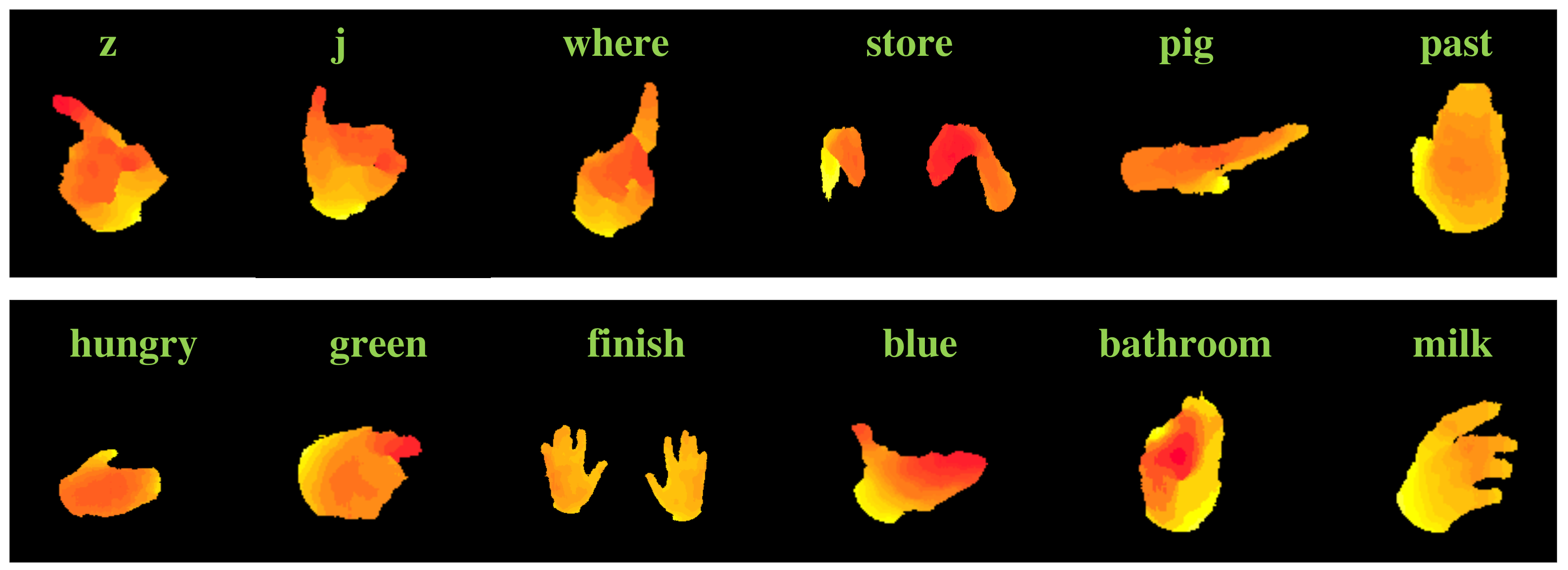}
\captionsetup{font={scriptsize}}
\vspace{-.5em}
\caption{Keyframes from the MSRGesture3D dataset
}\label{30}\vspace{-.5em}
\end{figure}

\begin{figure}[!htbp]
\centering
\includegraphics[width=.92\linewidth]{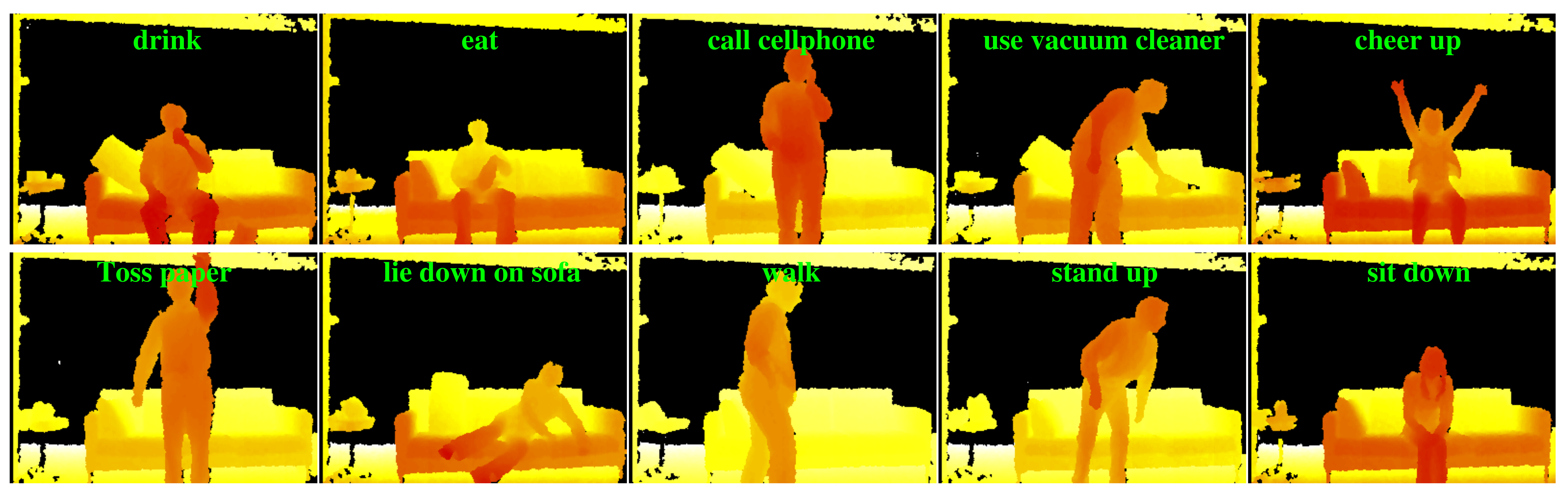}
\captionsetup{font={scriptsize}}
\vspace{-.5em}
\caption{Keyframes from the MSRDailyActivity3D dataset
}\label{31}\vspace{-0.5em}
\end{figure}

\begin{figure}[!htbp]
\centering
\includegraphics[width=.92\linewidth]{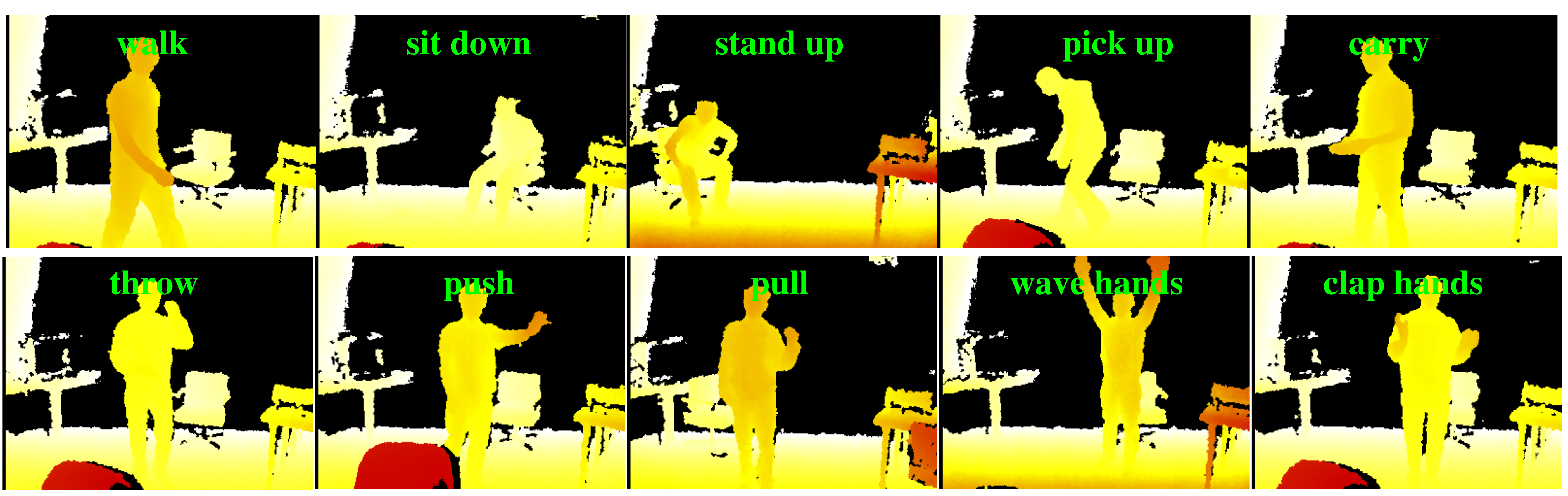}
\captionsetup{font={scriptsize}}
\vspace{-.5em}
\caption{Keyframes from the UTKinect dataset
}\label{32}
\end{figure}

\section{Experiments}
\subsection{Benchmark Datasets}
The MSRAction3D dataset \cite{li2010action} stands out as one of the most frequently used depth datasets for action recognition as reported in \cite{padilla2014discussion}.
It contains 20 actions and each action is performed two or three times by 10 subjects that are facing the depth camera, resulting in a total of 567 depth sequences in the dataset.
This is a challenging dataset for action recognition, since many actions in this dataset are highly similar (see Fig. \ref{28}). For example, actions such as ``drawX" and ``drawTick" share similar basic movements, with slight differences in the movements of one hand.
We adopt two widely used validation methods for the MSRAction3D dataset: ``cross-subject I" and ``cross-subject II".
In ``cross-subject I", we use cross-subject validation with subjects 1, 3, 5, 7, and 9 for training and subjects 2, 4, 6, 8, and 10 for testing \cite{li2010action}.
In ``cross-subject II", we divide the original dataset into three subsets, each consisting of eight actions. Then, a cross-subject validation scheme is used, with subjects 1, 3, 5, 7, and 9 for training and subjects 2, 4, 6, 8, and 10 for testing. The overall accuracy is calculated by taking the average over three subsets \cite{li2010action}.

\begin{figure*}[t]
\begin{center}
\includegraphics[width=1\linewidth]{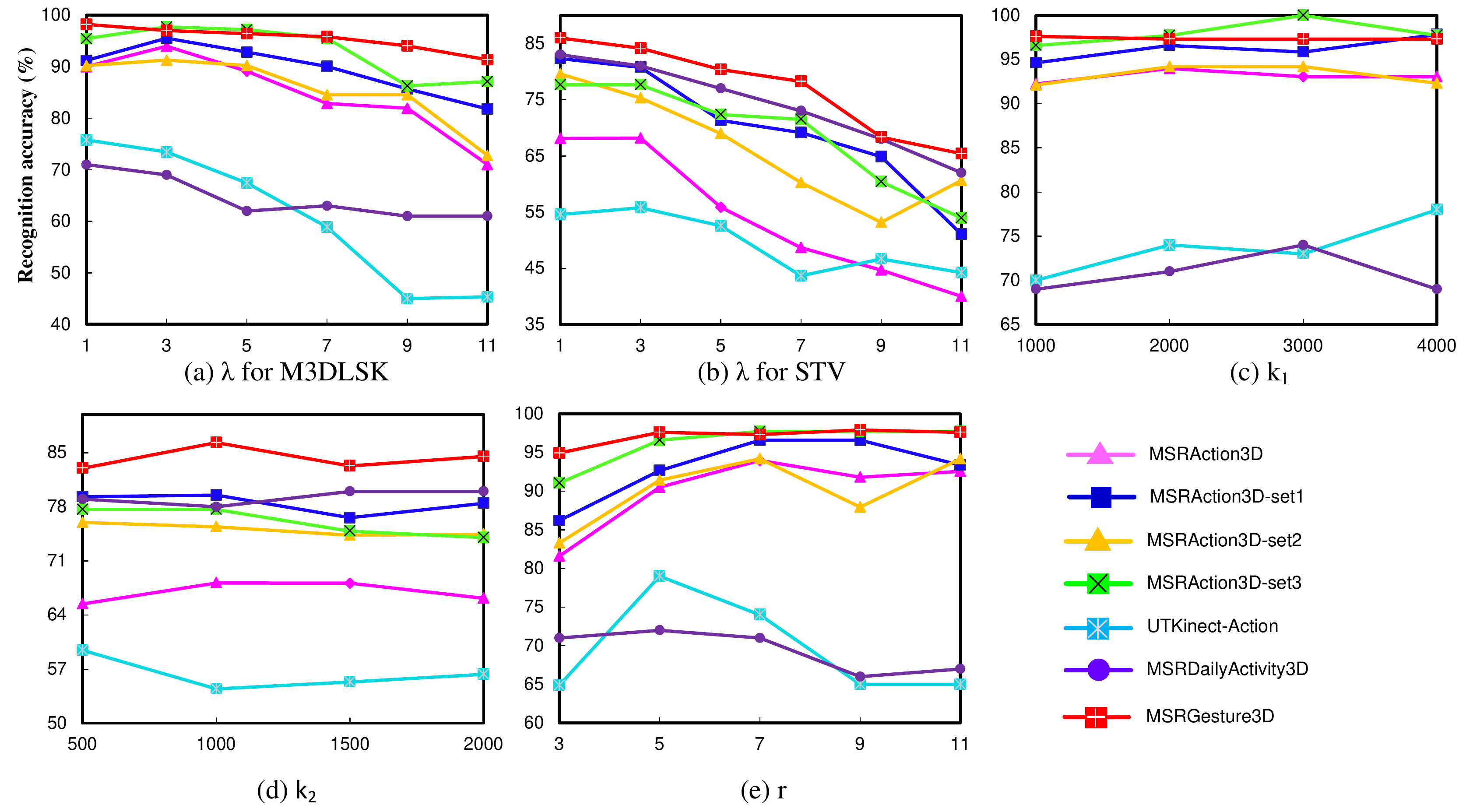}
\end{center}
\vspace{-1em}
\captionsetup{font={scriptsize}}
\caption{Evaluation of different parameters. The default values are set to $\lambda=3, k_1=2000, k_2=1000$, and $r=7$.}\label{33}\vspace{-.5em}
\end{figure*}

The MSRGesture3D dataset \cite{wang2012robust} is a popular dataset for hand gesture recognition \cite{padilla2014discussion}.
It contains 12 gestures that are defined by American Sign Language. Each action is performed two or three times by each subject, resulting in 336 depth sequences.
This is a challenging dataset for gesture recognition, since inter-class similarities among different types of gestures are observed (see Fig. \ref{30}).
For example, gestures such as ``milk" and ``hungry" are similar, since both actions involve the motion of bending the palm.
Additionally, self-occlusion is also a main issue for this dataset.
To ensure a fair comparison, we employ the leave-one-out cross-validation scheme described in \cite{wang2012robust}.

The MSRDailyActivity3D dataset \cite{wang2012mining} is a daily activity dataset, which contains 16 activities. Each action is performed by each subject in two different poses: ``sitting on sofa" and ``standing", resulting in 320 depth sequences.
This dataset contains background clutter and noise. Moreover, most of the actions contain human-object interactions, which are illustrated in Fig. \ref{31}.
We use cross-subject validation with subjects 1, 3, 5, 7, and 9 for training and subjects 2, 4, 6, 8, and 10 for testing \cite{wang2012mining}.

The UTKinect-Action dataset \cite{Xia2012View} consists of 200 sequences from 10 actions. Each action is performed two times by 10 subjects.
This dataset is designed to investigate variations in view point: right view, frontal view, right view and back view (see Fig. \ref{32}).
In addition, the background clutter and human-object interactions in some actions bring additional challenges for action recognition.
We use cross-subject validation with subjects 1, 2, 3, 4, and 5 for training and subjects 6, 7, 8, 9, and 10 for testing \cite{zhu2014evaluating}.
Note that this type of validation is more challenging than the leave-one-sequence-out scheme \cite{Xia2012View}.

\begin{figure*}[t]
\begin{center}
\includegraphics[width=.65\linewidth]{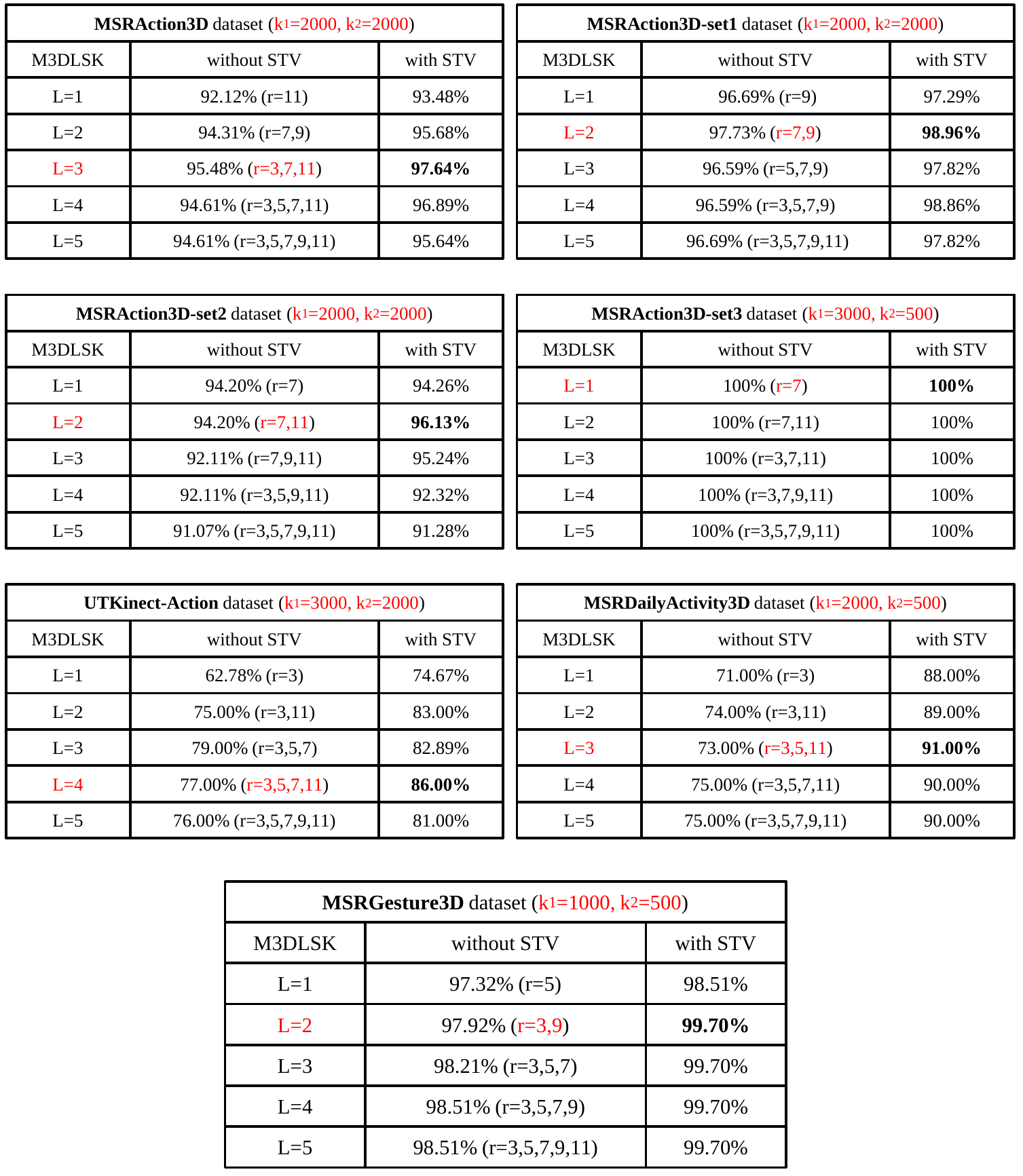}
\end{center}
\vspace{-1em}
\captionsetup{font={scriptsize}}
\caption{Selection of proper parameters, i.e., $k_1$, $k_2$, $r$, and $L$, to achieve the best performances on four benchmark datasets. Other parameters are set to default values.}\label{334}\vspace{-.5em}
\end{figure*}

\subsection{Parameter Settings}
Action recognition is conducted using a non-linear SVM with a homogeneous Chi2 kernel \cite{vedaldi2010vlfeat}, and the parameter $\gamma$ that determines the homogeneity degree of the kernel is set to 0.8. We choose the ``sdca" solver for SVM and use other default parameters provided in the VLFeat library \footnote{\url{http://www.vlfeat.org/applications/caltech-101-code.html}}.
In our BoVW model, we use vector quantization (VQ) for coding.
Since random initialization is involved in the clustering step of the BoVW model, we select the best performance over 10 runs.
For the representation of motion-based STIPs, there are three parameters: the size of the cuboid $r$, the sampling density $\lambda$ and the number of cluster centers $k_1$.
For the representation of shape-based STIPs, there are two parameters: the sampling density $\lambda$ and the number of cluster centers $k_2$.
We test the effect on the recognition accuracy of one parameter at a time, while keeping the other parameters at the default values. Bag of 3DLSK and Bag of STV representations are respectively used as basic methods to evaluate these parameters. The default values are set as
$\lambda=3, k_1=2000, k_2=1000$, and $r=7$. It is noted that the default values are used to evaluate 3DLSK and 3DLSK+STV.

In Fig. \ref{33} (a) and (b), the performances degrades when the sample density becomes sparse. This suggests that dense sampling should be used to obtain the motion-based and shape-based STIPs.
However, the computational cost is high when the STIPs are highly dense. Therefore, we set the value of $\lambda$ to $3$ in the following experiments, as a tradeoff between time cost and performance.
Fig. \ref{33} (c), (d) and (e) show that different settings of numbers of cluster centers and scales of cuboids are tuned for different datasets to achieve the best performance.
Therefore, we tune parameters $k_1,k_2$, and $r$ jointly by grid search in $[1000,2000,3000,4000]$, $[500,1000,1500,2000]$ and $[3,5,7,9,11]$, respectively.
For a given dataset, we apply the five-fold cross-validation method on the training set to tune these parameters.
As shown in Fig. \ref{334}, we select proper parameters to ensure the best performance of the M3DLSK+STV method and then use these parameters to evaluate M3DLSK and M3DLSK+STV.
Table \ref{my-label25253} shows selected values of parameter $\lambda, k_1, k_2, L, r$. Besides, we apply the five-fold cross-validation method on the training set to tune other parameters. Specifically, for the MSRAction3D dataset, $T_1$ is set to 0.8; $T_2$ is set to $0.01$ multiplied by the maximal value in the foreground region; the structuring element for the morphological operator is chosen as a ``disk'' structure with parameter 5; the radius of the cuboid $\Omega$ is set to 3.
For other datasets, we use same settings of these values, which work well on various datasets in following experiments.

\begin{table}[]
\centering
\scriptsize
\renewcommand\arraystretch{0.9}
\captionsetup{font={scriptsize}}
\caption{Selected parameters for different datasets}
\label{my-label25253}
\begin{tabular}{cccccc}
\hline
{Dataset}    & $\lambda$ &{$k_1$} & {$k_2$} & {$L$} & {$r$} \\ \hline
MSRAction3D        & 3             & 2000        & 2000        & 3          & 3,7,11     \\
MSRAction3D-set1   & 3             & 2000        & 2000        & 2          & 7,9        \\
MSRAction3D-set2    & 3            & 2000        & 2000        & 2          & 7,11       \\
MSRAction3D-set3    & 3           & 3000        & 500         & 1          & 7          \\
UTKinect-Action     & 3           & 3000        & 2000        & 4          & 3,5,7,11   \\
MSRDailyActivity3D   & 3         & 2000        & 500         & 3          & 3,5,11     \\
MSRGesture3D         & 3           & 1000        & 500         & 2          & 3,9        \\ \hline
\end{tabular}
\end{table}

\subsection{3DLSK+STV}

Some STIP detectors and descriptors are illustrated in Table \ref{35}.
According to the type of detector, we divide these methods into three categories: dense-sampling-based, skeleton-based and local-structure-based methods. The dense-sampling-based method commonly describes each pixel of the spatial-temporal space. Random sampling \cite{wang2012robust} and regular sampling \cite{Vieira2012STOP} are treated as special cases of dense sampling. In \cite{Vieira2012STOP}, the 4D space of a depth sequence is equally divided into grids, and then each cell is described. Without restricting the scale of the cells, Wang \textit{et al.} \cite{wang2012robust} randomly sampled cells at all possible locations and scales, which generated a larger number of local features.
Skeleton-based methods treat the 3D locations of skeleton joints as STIPs. The shortcomings of this type of method lie in three aspects. First, the number of skeleton joints is limited; thus, they cannot fully capture the motions. Second, skeleton joints can be accurately estimated only when action performers directly face the camera. In other words, these joints are usually not reliable when action performers are not in an upright position (e.g., sitting on a seat or lying on a bed). Moreover, partial occlusions seriously affect the accuracy of the skeleton extraction method. Third, the skeleton is not defined for the task of gesture recognition. Different from the above methods, local-structure-based methods detect motions by characterizing local structures. In \cite{laptev2005space}, Harris3D points are treated as STIPs, where the depth values change dramatically along arbitrary directions. Observing that true spatial-temporal corners are rare, Dollar et al. \cite{dollar2005behavior} proposed the Cuboid detector, which uses Gabor filters to find periodic motions. Recently, DSTIP \cite{xia2013spatio} and STK-D \cite{Rahmani2016Histogram} were designed specifically for depth sequences.

\begin{table}[t]
\centering
\scriptsize
\renewcommand\arraystretch{0.9}
\captionsetup{font={scriptsize}}
\caption{Evaluation of our detectors and basic descriptors}
\label{35}
\begin{tabular}{llcc}
\hline
{Local point detector} & {Local point descriptor} & {\begin{tabular}[c]{@{}c@{}}MSR-\\ Action3D\end{tabular}} & {\begin{tabular}[c]{@{}c@{}}MSR-\\ Gesture3D\end{tabular}} \\ \hline
\rowcolor[HTML]{FFFFC7}
Random sampling \cite{wang2012robust}    & ROP \cite{wang2012robust}     & {\color[HTML]{000000} 86.50\%}      & {\color[HTML]{000000} 88.50\%}    \\
\rowcolor[HTML]{FFFFC7}
Regular sampling \cite{Vieira2012STOP}    & STOP \cite{Vieira2012STOP}     & {\color[HTML]{000000} 84.80\%}      & -    \\
\rowcolor[HTML]{FFFFC7}
Dense sampling \cite{wang2009evaluation}    & HON4D+$D_{disc}$ \cite{oreifej2013hon4d}          & 88.89\%                & 92.45\%                                 \\
\rowcolor[HTML]{FFFFC7}
Dense sampling \cite{wang2009evaluation} & HON4D \cite{oreifej2013hon4d}        & 85.85\%                 & 87.29\%                                 \\
\rowcolor[HTML]{FFFFC7}
Dense sampling \cite{wang2009evaluation}   & Super Normal Vector \cite{yang2014super}              & 93.09\%                  & 94.74\%                                 \\
\rowcolor[HTML]{CBCEFB}
Skeleton \cite{Shotton2011Real}       & HOJ3D \cite{Xia2012View}               & 63.60\%                                 & -                                       \\
\rowcolor[HTML]{CBCEFB}
Norm. Skeleton \cite{wang2014learning}      & AE \cite{wang2014learning}       & 81.60\%                                 & -                                       \\
\rowcolor[HTML]{CBCEFB}
Norm. Skeleton \cite{wang2014learning}  & LARP \cite{Vemulapalli2014Human}    & 78.80\%                    & -                                       \\
\rowcolor[HTML]{FFCCC9}
Harris3D \cite{laptev2008learning}           & HOG3D \cite{klaser2008spatio}               & 81.43\%                                 & 85.23\%                                 \\
\rowcolor[HTML]{FFCCC9}
Cuboid \cite{dollar2005behavior}            & HOG/HOF \cite{laptev2008learning}             & 78.70\%                                 & -                                       \\
\rowcolor[HTML]{FFCCC9}
DSTIP \cite{xia2013spatio}              & DCSF \cite{xia2013spatio}                & 89.30\%                                 & -                                       \\
\rowcolor[HTML]{FFCCC9}
STK-D \cite{Rahmani2016Histogram}              & Local HOPC \cite{Rahmani2016Histogram}          & 86.50\%                                 & 94.70\%                                 \\  \hline
\rowcolor[HTML]{FFCCC9}
{\color[HTML]{FE0000}\textbf{Motion-based}}&{\color[HTML]{FE0000}\textbf{3DLSK}}               & 93.98\%                                 & 97.32\%                                 \\
\rowcolor[HTML]{FFCCC9}
{\color[HTML]{FE0000}\textbf{Shape-based}}  & {\color[HTML]{FE0000}\textbf{STV}}                 & 68.18\%                                 & 86.31\%                                 \\
\rowcolor[HTML]{FFCCC9}
{\color[HTML]{FE0000}\textbf{Motion-Shape-based}}          & {{\color[HTML]{FE0000}\textbf{3DLSK+STV}}}         & {\color[HTML]{FE0000} \textbf{95.36\%}} & {\color[HTML]{FE0000} \textbf{98.95\%}} \\ \hline
\end{tabular}\vspace{-0.5em}
\end{table}

\begin{table}[t]
\centering
\scriptsize
\captionsetup{font={scriptsize}}
\caption{\textcolor[rgb]{0,0,0}{Evaluation of STV. Parameters are set to default values: $k_1=2000,k_2=1000$.}}
\label{36}
\begin{tabular}{lccc}
\hline
{Method} &{MSRAction3D}           & {MSRGesture3D}           & {Dimension}        \\ \hline
3DLSK                             & 93.98\%                                & 97.32\%                         & $k_1$  \\
3DLSK+{\color[HTML]{FE0000}\textbf{STV}}                         & {\color[HTML]{FE0000} \textbf{95.36\%}}  & {\color[HTML]{FE0000} \textbf{98.95\%}} & $k_1+k_2$ \\
3DLSK+STP (two levels)             & 94.32\%                                 & 97.59\%                          & $k_1 \times 5$       \\
3DLSK+STP (three levels)             & 94.61\%                                 & 97.89\%                  & $k_1 \times 23$                 \\
3DLSK+STW              & 94.36\%                                 & 97.62\%                & $k_1$                 \\\hline
\end{tabular}
\end{table}

\begin{table}[t]
\centering
\scriptsize
\renewcommand\arraystretch{1}
\captionsetup{font={scriptsize}}
\caption{Evaluation of M3DLSK+STV on seven benchmark datasets}
\label{38}
\begin{tabular}{
>{\columncolor[HTML]{FFFFFF}}l
>{\columncolor[HTML]{FFFFFF}}r
>{\columncolor[HTML]{FFFFFF}}r
>{\columncolor[HTML]{FFFFFF}}r }
\hline
{3D Action Dataset} & {\color[HTML]{FE0000}\textbf{STV}} & \textbf{\begin{tabular}[c]{@{}c@{}}{\color[HTML]{FE0000}\textbf{M3DLSK}}\end{tabular}} & \textbf{\begin{tabular}[c]{@{}c@{}}{\color[HTML]{FE0000}\textbf{M3DLSK+STV}}\end{tabular}} \\ \hline
MSRAction3D & 66.18\% & 95.48\% & {\color[HTML]{FE0000} \textbf{97.64\%}} \\
MSRAction3D-Set1 & 78.47\% & 97.73\% & {\color[HTML]{FE0000} \textbf{98.96\%}} \\
MSRAction3D-Set2 & 77.46\% & 94.2\% & {\color[HTML]{FE0000} \textbf{96.13\%}} \\
MSRAction3D-Set3 & 77.65\% & {\color[HTML]{FE0000} \textbf{100\%}} & {\color[HTML]{FE0000} \textbf{100\%}} \\
UTKinect-Action & 56.33\% & 77\% & {\color[HTML]{FE0000} \textbf{86.00\%}} \\
MSRDailyActivity3D & 79\% & 73\% & {\color[HTML]{FE0000} \textbf{91.00\%}} \\
MSRGesture3D & 83\% & 97.92\% & {\color[HTML]{FE0000} \textbf{99.70\%}} \\ \hline
\end{tabular}
\end{table}

\begin{figure*}[t]
\centering
\includegraphics[width=1\linewidth]{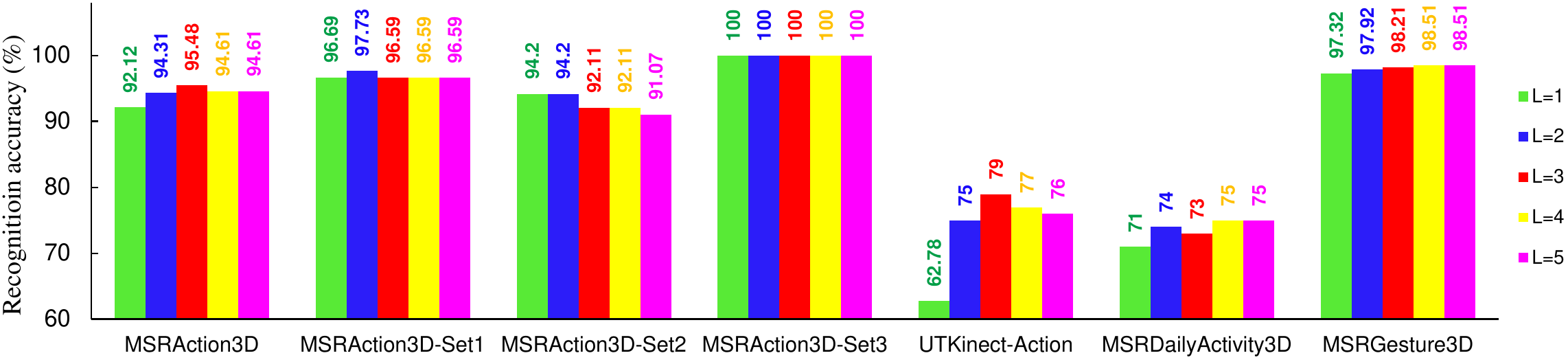}
\captionsetup{font={scriptsize}}
\vspace{-1.5em}
\caption{Evaluation of the M3DLSK descriptor with different numbers of scales
}\label{37}\vspace{-1.5em}
\end{figure*}

Along with the above detectors, various descriptors are provided. These descriptors can be roughly divided into three categories: traditional descriptors designed for RGB sequences, e.g., HOG3D \cite{klaser2008spatio} and HOG/HOF \cite{laptev2008learning}; skeleton-based descriptors, e.g., HOJ3D \cite{Xia2012View} and LARP \cite{Vemulapalli2014Human}; and depth-based descriptors, e.g., DCSF \cite{xia2013spatio}, Super Normal Vector \cite{yang2014super}, HON4D \cite{oreifej2013hon4d} and Local HOPC \cite{Rahmani2016Histogram}.
Among these descriptors, Super Normal Vector achieves best performances on both datasets.
Combining motion-based STIPs and 3DLSK, we achieve higher accuracy compared to all other methods. This confirms the effect of our STIP detection method and basic descriptor. Since 3DLSK and STV capture complementary information, their combination (3DLSK+STV) further improves the performance by $1.38\%$ on the MSRAction3D dataset and $1.63\%$ on the MSRGesture3D dataset.
Note that our method outperforms all dense sampling based methods, which verifies the importance of selecting motion-related or shape-related STIPs.

\textcolor[rgb]{0,0,0}{
To evaluate the power of STV in capturing the spatial-temporal distribution of STIPs, we further compare STV with two basic spatial-temporal information encoding methods: spatial-temporal pyramid (STP) and spatial-temporal weighting (STW). Three levels of the STP are set to $1\times1\times1$, $2\times2\times1$ and $3\times3\times2$, which achieve best performances for both datasets. Note that the dimension of the STP with two levels is $k_1 \times (1\times1\times1+2\times2\times1)$ and the dimension of the STP with three levels is $k_1 \times (1\times1\times1+2\times2\times1+3\times3\times2)$.
To implement the STW, we normalize the distance between an STIP and the origin of the coordinates to $[0,1]$, and then assign this value as the weight of the STIP. The Bag-of-Visual-Words model is used to characterize all STIPs as a weighted histogram. As shown in Table \ref{36}, our 3DLSK+STV achieves higher accuracies than 3DLSK+STP and 3DLSK+STW. What is more, the dimension of 3DLSK+STV is much lower than that of 3DLSK+STP, indicating lower time cost of classification.}

\begin{table}[t]
\renewcommand\arraystretch{0.96}
\centering
\scriptsize
\renewcommand\arraystretch{0.9}
\captionsetup{font={scriptsize}}
\caption{Comparison of recognition accuracy on the MSRAction3D dataset between our method and the previous approaches using ``cross-subject I" validation}\label{41}
\vspace{-0.6em}
\begin{tabular}{lcr}
\hline
         Methods                            &Accuracy  &   Details     \\
\hline
    \rowcolor[HTML]{CBCEFB}Actionlet Ensemble \cite{wang2014learning} &88.20\% &Wang \textsl{et al.} (2014) \\

    \rowcolor[HTML]{CBCEFB}Moving Pose \cite{zanfir2013moving} &91.70\% &Zanfir \textsl{et al.} (2013) \\
    \rowcolor[HTML]{CBCEFB}Hierarchical RNN \cite{Du2015Hierarchical} &94.49\% &Du \textsl{et al.} (2015) \\
    \rowcolor[HTML]{CBCEFB}JAS (Cosine)+MaxMin+$HOG^2$ \cite{Ohn2013Joint} &94.84\% &Ohn \textsl{et al.} (2013) \\
    \rowcolor[HTML]{CBCEFB}DL-GSGC+TPM \cite{luo2013group} &96.70\% & Luo \textsl{et al.} (2013) \\
    \rowcolor[HTML]{FFFC9E}Depth Gradients+RDF \cite{rahmani2014real} &88.82\% &Rahmani \textsl{et al.} (2014) \\
    \rowcolor[HTML]{FFFC9E}Skeleton+LOP+HON4D \cite{Shahroudy2015Multimodal} &93.10\% & Shahroudy \textsl{et al.} (2015) \\
    \rowcolor[HTML]{FFFC9E}Multi-fused features \cite{Jalal2017Robust} &93.30\% & Jalal \textsl{et al.} (2017) \\
    \rowcolor[HTML]{FFCCC9}Bag of 3D Points \cite{li2010action} &74.70\% &Li \textsl{et al.} (2010) \\
    \rowcolor[HTML]{FFCCC9}Motion Depth Surface \cite{Azary2013Grassmannian} &78.48\% &Azary \textsl{et al.} (2013) \\
    \rowcolor[HTML]{FFCCC9}STOP \cite{Vieira2012STOP} &84.80\% &Vieira \textsl{et al.} (2012) \\
    \rowcolor[HTML]{FFCCC9}Random Occupancy Pattern \cite{wang2012robust} &86.50\% &Wang \textsl{et al.} (2012) \\
    \rowcolor[HTML]{FFCCC9}STK-D+Local HOPC \cite{Rahmani2016Histogram} & 86.50\% & Rahmani \textsl{et al.} (2016) \\
    \rowcolor[HTML]{FFCCC9}LSTM \cite{veeriah2015differential} &87.78\% &Veeriah \textsl{et al.} (2015) \\
    \rowcolor[HTML]{FFCCC9}Depth Motion Maps \cite{yang2012recognizing} &88.73\% &Yang \textsl{et al.} (2012) \\
    \rowcolor[HTML]{FFCCC9}HON4D \cite{oreifej2013hon4d} &88.89\% &Oreifej \textsl{et al.} (2013) \\
    \rowcolor[HTML]{FFCCC9}DSTIP+DCSF \cite{xia2013spatio} &89.30\% &Xia \textsl{et al.} (2013) \\
    \rowcolor[HTML]{FFCCC9}H3DF \cite{Zhang2015Histogram} &89.45\% &Zhang \textsl{et al.} (2015) \\
    \rowcolor[HTML]{FFCCC9}LSGF \cite{Zhang2016Local} &90.76\% &Zhang \textsl{et al.} (2016) \\
    \rowcolor[HTML]{FFCCC9}HOG3D+LLC \cite{Rahmani2015Discriminative} &90.90\% &Rahmani \textsl{et al.} (2015) \\
    \rowcolor[HTML]{FFCCC9}Volumetric spatial feature \cite{Cho2015Volumetric} &91.30\% &Cho \textsl{et al.} (2015) \\
    \rowcolor[HTML]{FFCCC9}dRNN \cite{veeriah2015differential} &92.03\% &Veeriah \textsl{et al.} (2015) \\
    \rowcolor[HTML]{FFCCC9}Hierarchical 3D Kernel \cite{Kong2015Hierarchical} &92.73\% &Kong \textsl{et al.} (2015) \\
    \rowcolor[HTML]{FFCCC9}DMM-LBP-DF \cite{Chen2015Action} &93.00\% &Chen \textsl{et al.} (2015) \\
    \rowcolor[HTML]{FFCCC9}4DCov+Sparse Collab. \cite{Cirujeda20144DCov} & 93.01\% & Cirujeda \textsl{et al.} (2014) \\
    \rowcolor[HTML]{FFCCC9}Super Normal Vector \cite{Yang2016Super} &93.45\% &Yang \textsl{et al.} (2017) \\
    \rowcolor[HTML]{FFCCC9}Depth Context \cite{Liu2015Depth} &94.28\% &Liu \textsl{et al.} (2016) \\
    \rowcolor[HTML]{FFCCC9}Range-Sample \cite{Lu2014Range} &95.62\% & Lu \textsl{et al.} (2014) \\
    \rowcolor[HTML]{FFCCC9}Multi-scale E-GTIs \cite{liu20173d} &97.27\% & Liu \textsl{et al.} (2017) \\ \rowcolor[HTML]{FFCCC9}{{\color[HTML]{FE0000}\textbf{M3DLSK+STV}}} &\textcolor[rgb]{1,0,0}{\textbf{97.64\%}} & Three scales\\
\hline
\end{tabular}
\vspace{-0em}
\end{table}

\begin{figure}[t]
\begin{center}
\includegraphics[width=1\linewidth]{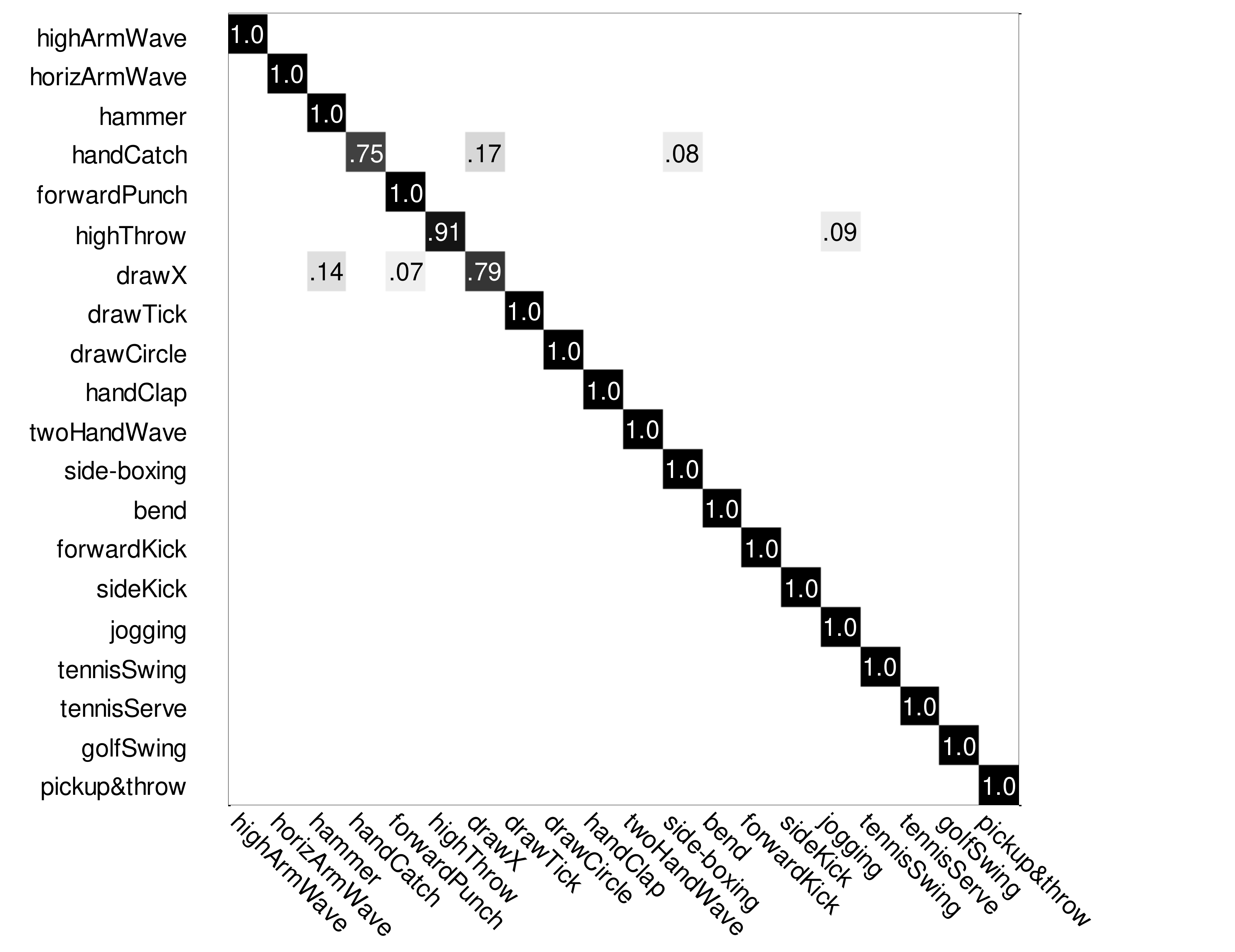}
\end{center}
\vspace{-1.0em}
\captionsetup{font={scriptsize}}
\caption{Confusion matrix on MSRAction3D dataset using ``cross-subject I" validation}\label{40}\vspace{-.0em}
\end{figure}

\subsection{M3DLSK+STV}

Fig. \ref{37} shows the performance of the M3DLSK descriptor with different numbers of scales. We treat 3DLSK as a specific case of M3DLSK by setting the value of $L$ to $1$. Generally, the accuracy first increases and then decreases when $L$ ranges from $1$ to $5$. The reason for the increase is that the multiple scales of 3DLSK provide more information about local structures than a single scale of 3DLSK. However, too many scales of 3DLSK may contain redundant information and will decrease the discriminative power, which explains the poor performance when $L$ is equal to $5$. For each dataset, we select the proper value of the scale by grid search for the grid $[1, 2, 3, 4, 5]$, where a five-fold cross-validation method is used on the training data.

Table \ref{38} evaluates the complementarity property between M3DLSK and STV.
As seen from this table, M3DLSK+STV outperforms M3DLSK on all datasets.
On the MSRDailyActivity3D dataset, M3DLSK+STV achieves an accuracy of $91.00\%$, which is $18\%$ higher than that of M3DLSK. Surprisingly, STV achieves an accuracy of $79\%$, which is $6\%$ higher than that of M3DLSK. These improvements show that STV is more effective than M3DLSK in capturing the shape cues of human-object interactions.
On the UTKinect-Action dataset, M3DLSK+STV achieves an accuracy of $86\%$, which is $9\%$ higher than that of M3DLSK. This result also verifies the complementarity property between M3DLSK and STV.

\setlength{\tabcolsep}{2pt}
\begin{table}[t]
\renewcommand\arraystretch{0.9}
\centering
\scriptsize
\captionsetup{font={scriptsize}}
\caption{Comparison of recognition accuracy on the MSRAction3D dataset between our method and the previous approaches using ``cross-subject II" validation}\label{42}
\vspace{-0.6em}
\begin{tabular}{lccccc}
\hline
         Methods&Set1 &Set2 &Set3 &Overall &Details     \\
\hline
   \rowcolor[HTML]{CBCEFB}HOJ3D \cite{Xia2012View} &87.98\%&85.48\%&63.46\%&78.97\%&2012 \\
   \rowcolor[HTML]{CBCEFB}Covariance descriptor \cite{hussein2013human}&88.04\%&89.29\%&94.29\%&90.53\%&2013 \\
   \rowcolor[HTML]{CBCEFB}Skeletal quads \cite{evangelidis2014skeletal}&88.39\%&86.61\%&94.59\%&89.86\%&2014 \\
   \rowcolor[HTML]{CBCEFB}LARP \cite{Vemulapalli2014Human}&95.29\%&83.87\%&98.22\%&92.46\%&2014 \\
   \rowcolor[HTML]{CBCEFB}Hierarchical RNN \cite{Du2015Hierarchical} &93.33\%&94.64\%&95.50\%&94.49\%&2015 \\
   \rowcolor[HTML]{FFFC9E}Multi-fused features \cite{Jalal2017Robust} &90.80\%&93.40\%&95.70\% & 93.30\% &2017 \\
   \rowcolor[HTML]{FFCCC9}Bag of 3D Points \cite{li2010action} &72.90\%&71.90\%&79.20\%&74.70\%&2010 \\
   \rowcolor[HTML]{FFCCC9}3D$^2$CNN \cite{Liu20163D} &86.79\%&76.11\%&89.29\% & 84.07\%&2016 \\
   \rowcolor[HTML]{FFCCC9}STOP \cite{Vieira2012STOP} &84.70\%&81.30\%&88.40\%&84.80\%&2012 \\
   \rowcolor[HTML]{FFCCC9}{{\color[HTML]{FE0000}\textbf{M3DLSK+STV}}}&\textcolor[rgb]{1,0,0}{\textbf{98.96\%}}&\textcolor[rgb]{1,0,0}{\textbf{96.13\%}}&\textcolor[rgb]{1,0,0}{\textbf{100\%}}&\textcolor[rgb]{1,0,0}{\textbf{98.36\%}} & -\\
\hline
\end{tabular}\vspace{-0em}

\end{table}
\begin{figure}[t]
\begin{center}
\includegraphics[width=1\linewidth]{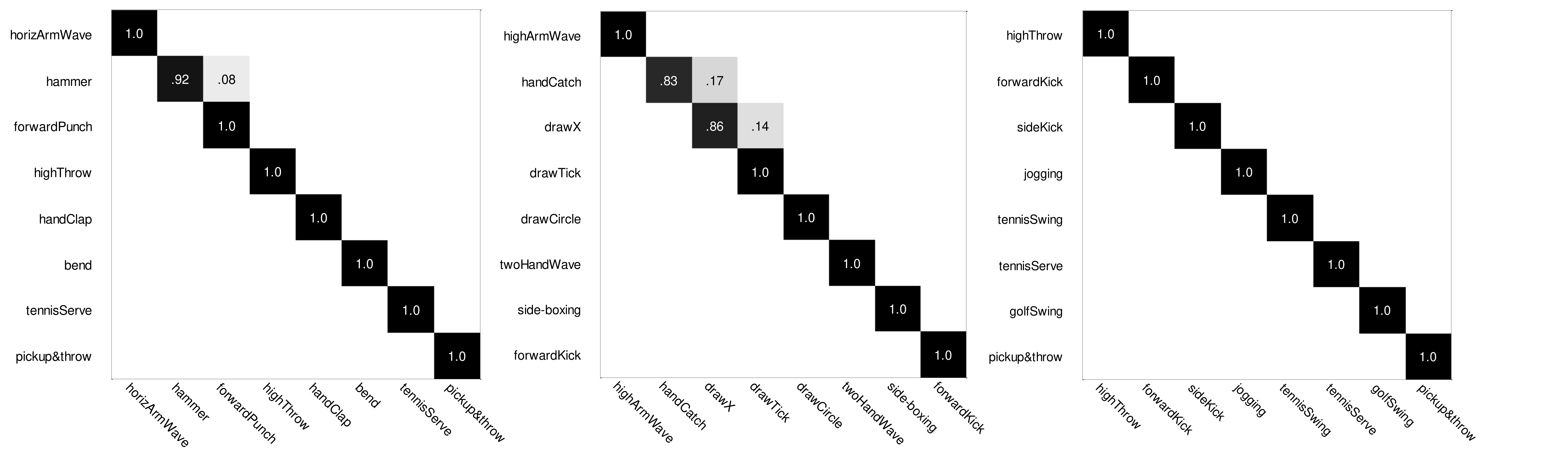}
\end{center}
\vspace{-1em}
\captionsetup{font={scriptsize}}
\caption{Confusion matrix on MSRAction3D dataset using ``cross-subject II" validation}\label{39}\vspace{0em}
\end{figure}

\subsection{Comparison with Existing Approaches}
\subsubsection{MSRAction3D dataset}
In Table \ref{41}, we compare our method with the previous approaches on the MSRAction3D dataset using ``cross-subject I" validation. We divide these approaches into three categories: skeleton-based (colored in blue), depth-based (colored in red) and hybrid-based (colored in yellow) methods. Note that Actionlet Ensemble \cite{wang2014learning} uses both skeleton joints and depth images. We classify this type of method as a skeleton-based method. Since only depth information is used in our method, we mainly compare our approach with depth-based methods.

Skeleton-based methods such as \cite{Ohn2013Joint} and \cite{luo2013group} achieve high accuracies. In \cite{Ohn2013Joint}, the depth value is described by the $HOG^2$ feature, and the spatial-temporal distributions of skeleton joints are encoded by \textcolor[rgb]{0,0,0}{Joint Angles Similarities} (JAS) feature. In \cite{luo2013group}, an improved version of the sparse coding method is proposed, which dramatically improves the performance of the skeleton-based feature. However, skeleton data may not be reliable when the subject is not in an upright position. Moreover, the skeleton data are not available from depth sequences that contain partial human bodies, e.g., hands, arms and legs. Therefore, applications of skeleton-based methods are limited.

\setlength{\tabcolsep}{2pt}
\begin{table}[t]
\renewcommand\arraystretch{0.9}
\centering
\scriptsize
\captionsetup{font={scriptsize}}
\caption{Comparison of recognition accuracy on the MSRGesture3D dataset between our method and the previous approaches}\label{43}
\vspace{-0.6em}
\footnotesize
\begin{tabular}{lcr}
\hline
         Methods                            &Accuracy  &   Details     \\
\hline

    \rowcolor[HTML]{FFCCC9}Motion Depth Surface \cite{Azary2013Grassmannian} &85.42\% &Azary \textsl{et al.} (2013) \\
    \rowcolor[HTML]{FFCCC9}Random Occupancy Pattern \cite{wang2012robust} &88.50\% &{Wang \textsl{et al.} (2012)} \\
    \rowcolor[HTML]{FFCCC9}HON4D \cite{oreifej2013hon4d} &92.45\% &Oreifej \textsl{et al.} (2013) \\
    \rowcolor[HTML]{FFCCC9}4DCov+Sparse Collab. \cite{Cirujeda20144DCov} & 92.89\% & Cirujeda \textsl{et al.} (2014) \\
    \rowcolor[HTML]{FFCCC9}HOG3D+LLC \cite{Rahmani2015Discriminative} &94.10\% &Rahmani \textsl{et al.} (2015) \\
    \rowcolor[HTML]{FFCCC9}DMM-LBP-DF \cite{Chen2015Action} &94.60\% &Chen \textsl{et al.} (2015) \\
    \rowcolor[HTML]{FFCCC9}3DHoT-MBC \cite{zhang2017action} &94.70\% &Zhang \textsl{et al.} (2017) \\
    \rowcolor[HTML]{FFCCC9}Super Normal Vector \cite{Yang2016Super} &94.74\% &Yang \textsl{et al.} (2016) \\
    \rowcolor[HTML]{FFCCC9}H3DF \cite{Zhang2015Histogram}           &95.00\% &Zhang \textsl{et al.} (2015) \\
    \rowcolor[HTML]{FFCCC9}Depth Gradients+RDF \cite{rahmani2014real} &95.29\% &Rahmani \textsl{et al.} (2014) \\
    \rowcolor[HTML]{FFCCC9}Hierarchical 3D Kernel \cite{Kong2015Hierarchical} &95.66\% &Kong \textsl{et al.} (2015) \\
    \rowcolor[HTML]{FFCCC9}STK-D+Local HOPC \cite{Rahmani2016Histogram} &96.23\% &Hossein \textsl{et al.} (2016) \\
    \rowcolor[HTML]{FFCCC9}{{\color[HTML]{FE0000}\textbf{M3DLSK+STV}}} &\textcolor[rgb]{1,0,0}{\textbf{99.70\%}} & Two scales\\
\hline
\end{tabular} \vspace{-0em}
\end{table}

Depth-based methods such as HON4D \cite{oreifej2013hon4d} and DSTIP \cite{xia2013spatio} do not perform well; however, Depth Context \cite{Liu2015Depth} and Range-Sample \cite{Lu2014Range} achieve high accuracies. Our M3DLSK+STV achieves an accuracy of $97.64\%$, which outperforms all depth-based methods. The main advantage of our method lies in the combination of both local and global spatial-temporal structures. The confusion matrix of our proposed M3DLSK+STV method is shown in Fig. \ref{40}. The actions ``handCatch" and ``drawx" have maximum confusion with each other because both actions share the motion ``raise a hand to one side of the human body and then bring it down". Similarly, the actions ``hammer" and ``drawX" have high confusion because they both contain similar motions and are of similar appearance.

\setlength{\tabcolsep}{2pt}
\begin{table}[t]
\renewcommand\arraystretch{0.9}
\centering
\scriptsize
\captionsetup{font={scriptsize}}
\caption{Comparison of recognition accuracy on the UTKincet-Action dataset between our method and the previous approaches}\label{44}
\vspace{-0.6em}
\footnotesize
\begin{tabular}{lcr}
\hline
         Methods                            &Accuracy  &   Details     \\
\hline
    \rowcolor[HTML]{CBCEFB}Skeleton Joint Features \cite{Zhao2012Combing} &87.90\% &Zhao \textsl{et al.} (2012) \\
    \rowcolor[HTML]{FFFC9E}Combined features with RFs \cite{Zhao2012Combing} &91.90\% &Zhao \textsl{et al.} (2012) \\
    \rowcolor[HTML]{FFCCC9}STIPs (Cuboids+HOG/HOF) \cite{dollar2005behavior} &65.00\% & Dollar \textsl{et al.} (2005) \\
    \rowcolor[HTML]{FFCCC9}STIPs (Harris3D+HOG3D) \cite{klaser2008spatio} &80.80\% &Klaser \textsl{et al.} (2008) \\
    \rowcolor[HTML]{FFCCC9}3D$^2$CNN \cite{Liu20163D} &82.00\% &Liu \textsl{et al.} (2016) \\
    \rowcolor[HTML]{FFCCC9}DSTIP+DCSF \cite{xia2013spatio}                & 85.80\% &Xia \textsl{et al.} (2013) \\
    \rowcolor[HTML]{FFCCC9}WHDMM+3ConvNets \cite{Wang2015Action}                & 90.91\% &Wang \textsl{et al.} (2015) \\
    \rowcolor[HTML]{FFCCC9}{{\color[HTML]{FE0000}\textbf{M3DLSK+STV}}} &\textcolor[rgb]{1,0,0}{\textbf{86.00\%}} & Four scales\\

\hline
\end{tabular} \vspace{0em}
\end{table}

In Table \ref{42}, we compare our method with the previous approaches on the MSRAction3D dataset using ``cross-subject II" validation. Bag of 3D Points \cite{li2010action} and STOP \cite{Vieira2012STOP} are most competitive with our work. In \cite{li2010action}, each STIP is treated as a 3D point, and the spatial-temporal distribution of the 3D points is encoded by an Action Graph model. In \cite{Vieira2012STOP}, each STIP is described by a space-time occupancy pattern. Both methods do not work well, since either the local structure or the global structure is ignored. Our method combines both cues and, therefore, achieves higher accuracy than existing depth-based methods. Hierarchical RNN \cite{Du2015Hierarchical} stands out as the state-of-the-art skeleton-based method. An improvement of $3.87\%$ is achieved over this method, which verifies that depth data can provide richer information than skeleton data. On MSRAction3D-Set3, we achieve $100\%$ accuracy, which verifies the efficacy of our method. The confusion matrix of our proposed M3DLSK+STV method is shown in Fig. \ref{39}.

\subsubsection{MSRGesture3D dataset}
In Table \ref{43}, we test the performance of our method on a gesture recognition task using a widely used MSRGesture3D dataset.
Since skeleton data are not available for this task, we only compare our method with depth-based methods.
As a local descriptor, HON4D \cite{oreifej2013hon4d} achieves an accuracy of $92.45\%$.
As a global descriptor, DMM-LBP-DF \cite{Chen2015Action} achieves an accuracy of $94.60\%$.
Our method outperforms both types of descriptors, since we jointly encode both local and global information.
An improvement of $3.47\%$ is achieved over a recent approach \cite{Rahmani2016Histogram}, which also shows the efficiency of our method.

\subsubsection{UTKinect-Action dataset}
In Table \ref{44}, we test the performance of our method against viewpoint changes using a benchmark: the UTKinect-Action dataset.
Compared to depth-based methods, skeleton-based methods are more robust to variations in viewpoint because more geometric information is obtained by skeleton joints and the positions of joints remain unchanged when the viewpoint changes.
Using skeleton data, an accuracy of $87.90\%$ is achieved using basic skeleton joint features \cite{Zhao2012Combing}.
Combining both skeleton and depth data with random forests, an accuracy of $91.90\%$ is achieved \cite{Zhao2012Combing}.
We compare our method with depth-based methods, since we do not need skeleton data, which are sensitive to partial occlusions and poses of human bodies.
Using depth data, the traditional STIP descriptor HOG3D achieves an accuracy of $80.8\%$ \cite{klaser2008spatio}.
Specifically, designed for depth data, our method achieves an accuracy of $86.00\%$, which is competitive with \cite{xia2013spatio}.
\textcolor[rgb]{0,0,0}{Weighted hierarchical depth motion maps (WHDMM) + three-channel deep convolutional neural networks (3ConvNets) \cite{Wang2015Action} achieves higher accuracy than our method, since a large number of depth sequences are synthesized to mimic different viewpoints. Therefore, WHDMM+3ConvNets can handle variations in viewpoint. However, extra training time is needed for fine-tuning neural networks.}

\setlength{\tabcolsep}{2pt}
\begin{table}[t]
\renewcommand\arraystretch{0.9}
\centering
\scriptsize
\captionsetup{font={scriptsize}}
\caption{Comparison of recognition accuracy on the MSRDailyActivity3D dataset between our method and the previous approaches}\label{45}
\vspace{-0.6em}
\footnotesize
\begin{tabular}{lcr}
\hline
         Methods                            &Accuracy  &   Details     \\
\hline
    \rowcolor[HTML]{CBCEFB}Joint position feature \cite{wang2014learning} &68.00\% &Wang \textsl{et al.} (2014) \\
    \rowcolor[HTML]{CBCEFB}Actionlet Ensemble \cite{wang2014learning} &86.00\% &Wang \textsl{et al.} (2014) \\
    \rowcolor[HTML]{CBCEFB}AND/OR Patterns on FTP features \cite{Weng2015Efficient} &86.88\% & Weng \textsl{et al.} (2015)\\
    \rowcolor[HTML]{CBCEFB}DSTIP+DCSF+Joint \cite{xia2013spatio}                & 88.20\% &Xia \textsl{et al.} (2013) \\
    \rowcolor[HTML]{CBCEFB}DL-GSGC+TPM \cite{luo2013group} &95.00\% & Luo \textsl{et al.} (2013) \\
    \rowcolor[HTML]{FFFC9E}Depth Gradients+RDF \cite{rahmani2014real} &81.25\% &Rahmani \textsl{et al.} (2014) \\
    \rowcolor[HTML]{FFFC9E}Skeleton+LOP+HON4D \cite{Shahroudy2015Multimodal} &91.25\% & Shahroudy \textsl{et al.} (2015) \\
    \rowcolor[HTML]{FFFC9E}Multi-fused features \cite{Jalal2017Robust} &94.10\% & Jalal \textsl{et al.} (2017) \\
    \rowcolor[HTML]{FFCCC9}LOP feature \cite{wang2014learning} &42.50\% &Wang \textsl{et al.} (2014) \\
    \rowcolor[HTML]{FFCCC9}STIPs (Harris3D+HOG3D) \cite{klaser2008spatio} &60.60\% &Klaser \textsl{et al.} (2008) \\
    \rowcolor[HTML]{FFCCC9}Random Occupancy Pattern \cite{wang2012robust} &64.00\% &Wang \textsl{et al.} (2012) \\
    \rowcolor[HTML]{FFCCC9}STIPs (Cuboids+HOG/HOF) \cite{dollar2005behavior} &70.60\% & Dollar \textsl{et al.} (2005) \\
    \rowcolor[HTML]{FFCCC9}HON4D \cite{oreifej2013hon4d} &80.00\% &Oreifej \textsl{et al.} (2013) \\
    \rowcolor[HTML]{FFCCC9}DSTIP+DCSF \cite{xia2013spatio}                & 83.60\% &Xia \textsl{et al.} (2013) \\
    \rowcolor[HTML]{FFCCC9}WHDMM+3ConvNets \cite{Wang2015Action}                & 85.00\% &Wang \textsl{et al.} (2015) \\
    \rowcolor[HTML]{FFCCC9}LSGF \cite{Zhang2016Local} &85.38\% &Zhang \textsl{et al.} (2016) \\
    \rowcolor[HTML]{FFCCC9}Super Normal Vector \cite{Yang2016Super} &86.25\% &Yang \textsl{et al.} (2016) \\
    \rowcolor[HTML]{FFCCC9}Volumetric spatial feature \cite{Cho2015Volumetric} &89.70\% &Cho \textsl{et al.} (2015) \\
    \rowcolor[HTML]{FFCCC9}{\color[HTML]{FE0000}\textbf{{M3DLSK+STV}}} &\textcolor[rgb]{1,0,0}{\textbf{91.00\%}} & Three scales\\
\hline
\end{tabular} \vspace{-0em}
\end{table}

\subsubsection{MSRDailyActivity3D dataset}
In Table \ref{45}, we test the performance of our method on human-object interactions using a benchmark: the MSRDailyActivity3D dataset.
Our method achieves an improvement of $7.40\%$ over DSTIP+DCSF \cite{xia2013spatio}, which shows the effectiveness of describing shapes by the STV descriptor. Our method even outperforms DSTIP+DCSF+Joint \cite{xia2013spatio}, which uses both depth and skeleton features. This result shows the efficacy of our method in characterizing depth data.
In \cite{luo2013group}, DL-GSGC+TPM achieves a state-of-the-art result of $95.00\%$, which outperforms all other methods.
We argue that the main improvement of \cite{luo2013group} lies in the use of an improved sparse coding method for action classification.
In our work, we mainly focus on the feature extraction and description steps and use a common SVM classifier for classification.
It is worth mentioning that our method can effectively handle partial occlusions, which could dramatically affect the performance of skeleton-based methods, e.g., \cite{luo2013group}.
\textcolor[rgb]{0,0,0}{Since original frames (including background clutter) are directly used as inputs, WHDMM+3ConvNets \cite{Wang2015Action} achieves lower accuracy than our method, which verifies the efficiency of extracting interest motions and shapes from background clutter.}

\begin{figure}[t]
\centering
\includegraphics[width=1\linewidth]{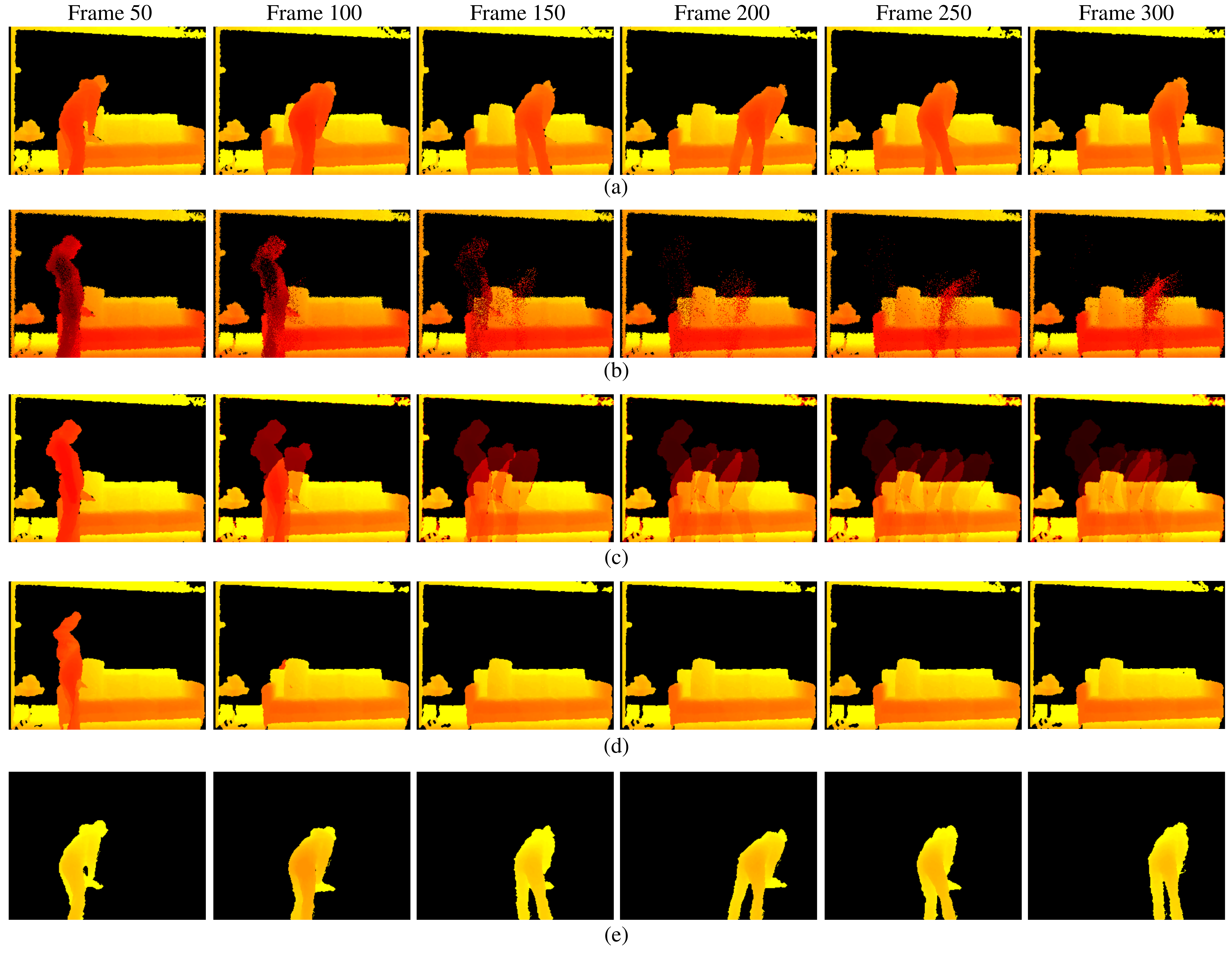}
\captionsetup{font={scriptsize}}
\vspace{-1.5em}
\caption{Background modeling method comparison. (a) A depth sequence; (b) Background formed by the ViBe method; (c) Background formed by the Mean filter method; (d) Background formed by our method; (e) Foreground extracted by our method.
}\label{34}\vspace{-0em}
\end{figure}

\subsection{Robustness}
\subsubsection{Robustness to background clutter}
Many background modeling methods have been developed for RGB data \cite{Bouwmans2014Traditional}.
However, the background modeling methods that use only depth data are limited.
Thereafter, we propose a new background modeling method and compare it with two other background modeling methods: ViBe \cite{Barnich2011ViBe} and Mean filter \cite{Benezeth2008Review}, which are widely applied on RGB data.
A depth sequence of the action ``use vacuum cleaner" from the MSRDailyActivity3D dataset (see Fig. \ref{34} (a)) is used for illustration.
\textcolor[rgb]{0,0,0}{Background formed by ViBe and Mean filter are shown in Fig. \ref{34} (b) and (c), where portions of the foreground are merged into the background.}
Since our method ignores the effect of foreground by considering the depth values, we can obtain clean background, as shown in Fig. \ref{34} (d). Using the background model in the last frame of Fig. \ref{34} (d), we obtain the foreground shown in Fig. \ref{34} (e), where the details of the human body and the object (i.e., the vacuum cleaner) are preserved and all fragments generated by noise are removed.
Examples of the cropped MSRDailyActivity3D and UTKinect-Action datasets can be found on the webpage \footnote{\url{https://github.com/liumengyuan/depthDatasets.git}}.

\begin{figure}[t]
\centering
\includegraphics[width=1\linewidth]{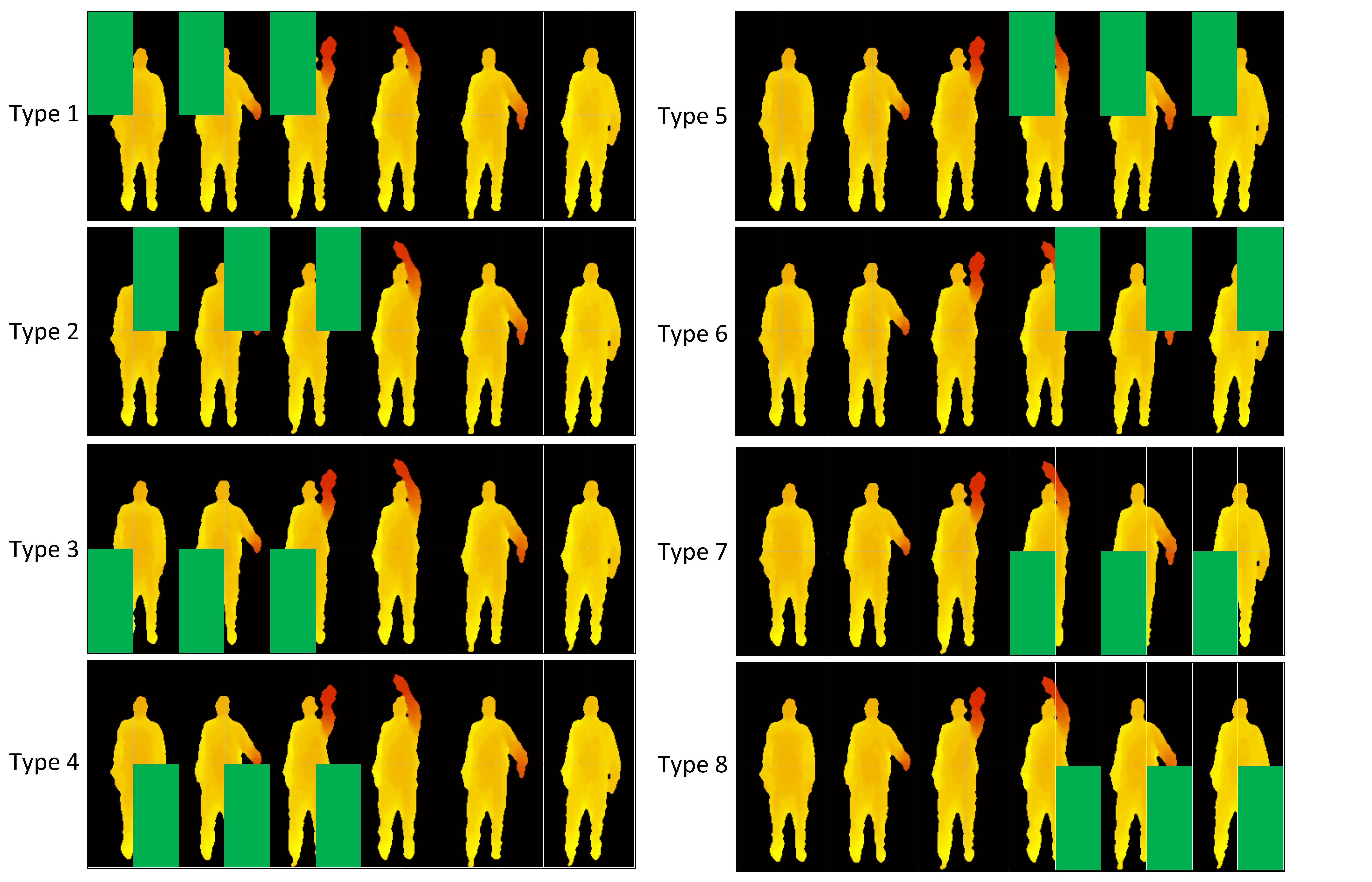}\\
\captionsetup{font={scriptsize}}\vspace{-0.5em}
\caption{\textcolor[rgb]{0,0,0}{Eight types of partial occlusions}}\label{359}\vspace{-.0em}
\end{figure}

\setlength{\tabcolsep}{2pt}
\begin{table}[t]\footnotesize
\centering
\captionsetup{font={scriptsize}}
\vspace{-.5em}
\caption{\textcolor[rgb]{0,0,0}{Evaluation of the robustness of different methods to partial occlusions}}\label{16}
\begin{tabular}{c|c|c|c}
\hline
         Occlusion     &ROP \cite{wang2012robust} &ROP+sparse coding \cite{wang2012robust}  &{{\color[HTML]{FE0000}\textbf{3DLSK+STV}}}     \\
\hline
none &85.92\% &86.20\% &\textcolor[rgb]{1,0,0}{\textbf{95.36\%}} \\
Type 1&83.05\% &86.17\% &\textcolor[rgb]{1,0,0}{\textbf{92.31\%}} \\
Type 2&84.18\% &86.50\% &\textcolor[rgb]{1,0,0}{\textbf{93.28\%}} \\
Type 3&78.76\% &80.09\% &\textcolor[rgb]{1,0,0}{\textbf{87.34\%}} \\
Type 4&82.12\% &85.49\% &\textcolor[rgb]{1,0,0}{\textbf{88.60\%}} \\
Type 5&84.48\% &87.51\% &\textcolor[rgb]{1,0,0}{\textbf{94.82\%}} \\
Type 6&82.46\% &87.51\% &\textcolor[rgb]{1,0,0}{\textbf{94.27\%}} \\
Type 7&80.10\% &83.80\% &\textcolor[rgb]{1,0,0}{\textbf{92.59\%}} \\
Type 8&85.83\% &86.83\% &\textcolor[rgb]{1,0,0}{\textbf{93.42\%}} \\
\hline
\end{tabular}
\end{table}

\begin{figure}[t]
\centering
\includegraphics[width=1\linewidth]{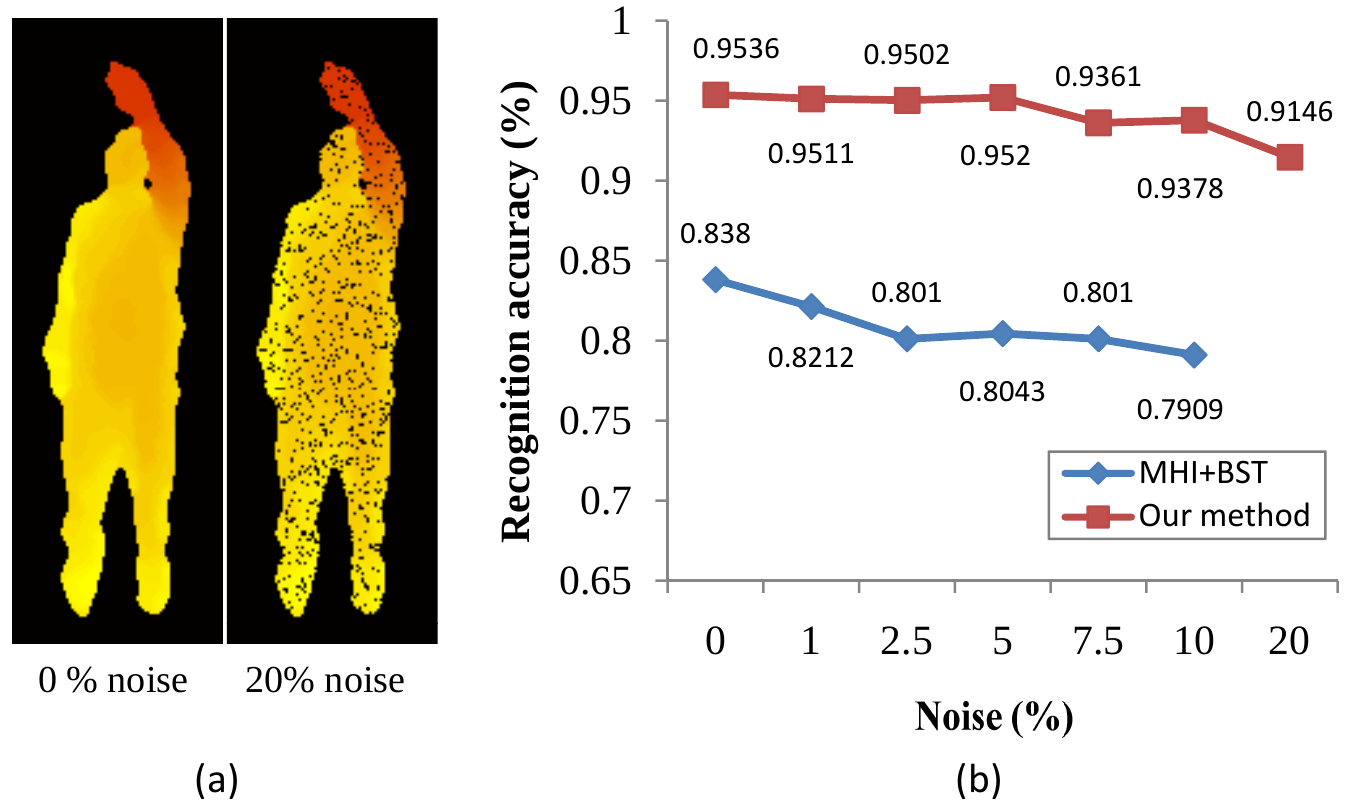}\\
\captionsetup{font={scriptsize}}\vspace{-0.5em}
\caption{\textcolor[rgb]{0,0,0}{(a) Original image and image affected by pepper noise. (b) Recognition
results on the MSRAction3D dataset with increasing percentages of
pepper noise: $0\%$, $1\%$, $2.5\%$, $5\%$, $7.5\%$, $10\%$ and $20\%$.}}\label{355}\vspace{-1em}
\end{figure}

\begin{figure}[t]
\begin{center}
\includegraphics[width=1\linewidth]{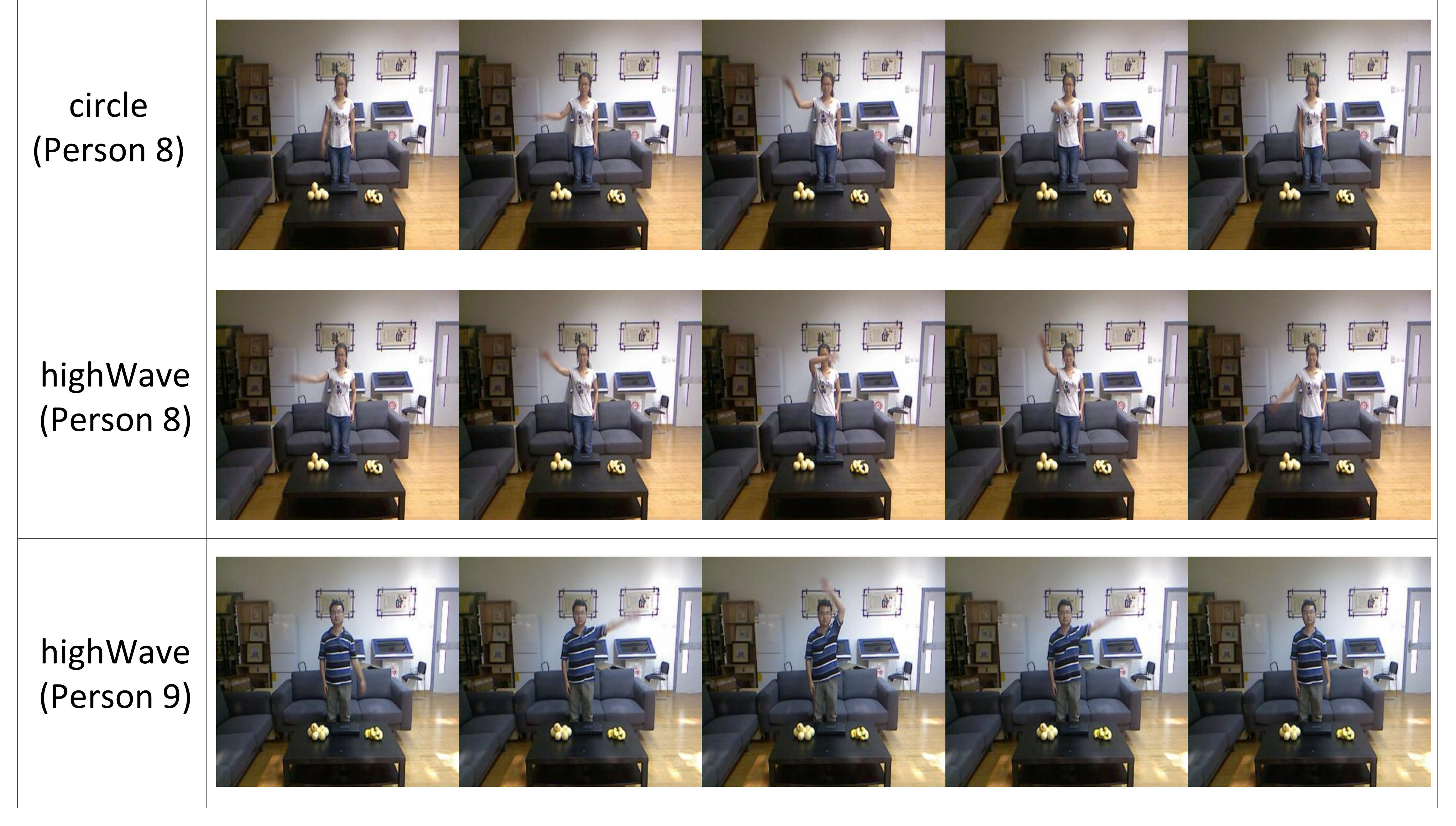}
\end{center}
\vspace{-1em}
\captionsetup{font={scriptsize}}
\caption{\textcolor[rgb]{0,0,0}{Inter-class similarities (``circle" and ``highWave" contain similar movements) and intra-class dissimilarities (``highWave" performed by person 8 and person 9 looks different) in the SmartHome dataset.}}\label{46}\vspace{-1em}
\end{figure}

\begin{figure}[!htbp]
\begin{center}
\includegraphics[width=1\linewidth]{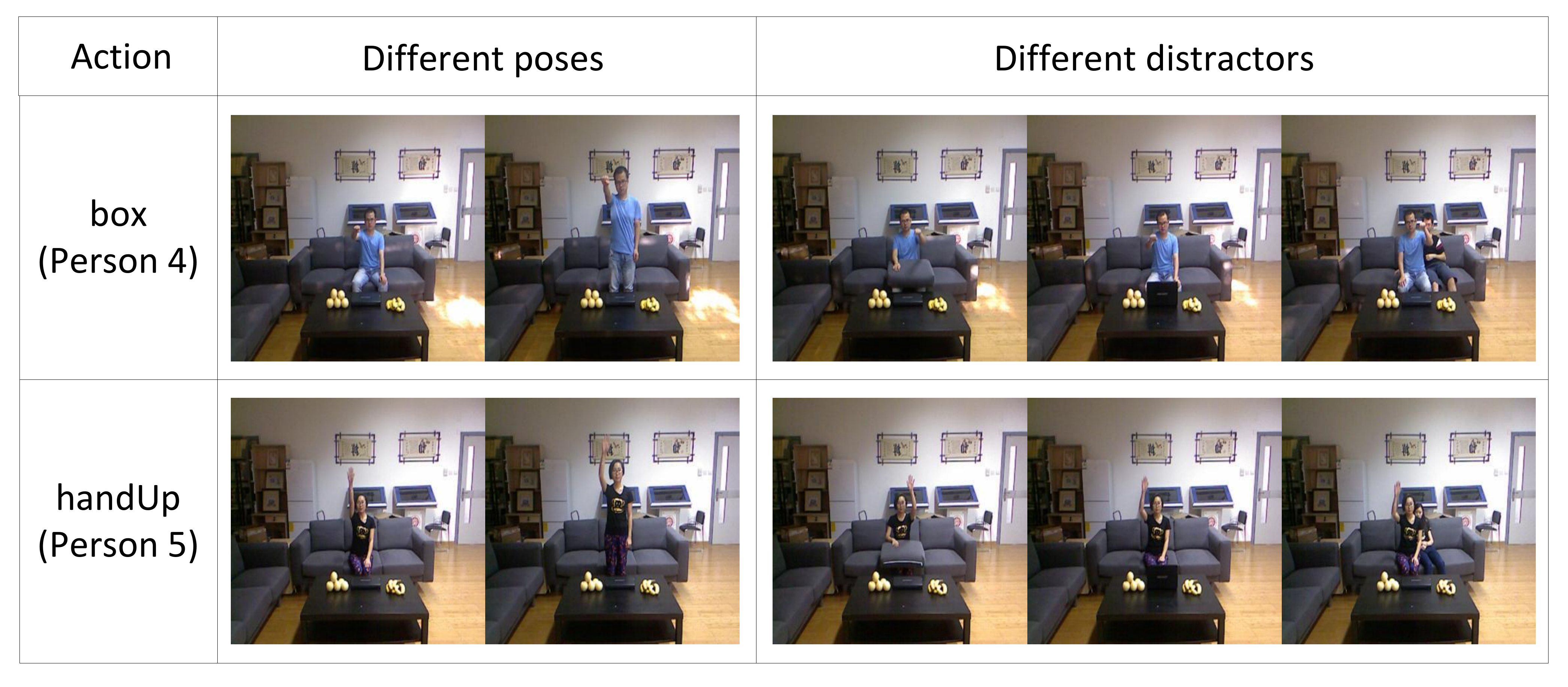}
\end{center}
\vspace{-1em}
\captionsetup{font={scriptsize}}
\caption{\textcolor[rgb]{0,0,0}{Different poses and distractors in the SmartHome dataset}}\label{47}\vspace{0em}
\end{figure}

\begin{figure}[t]
\begin{center}
\includegraphics[width=1\linewidth]{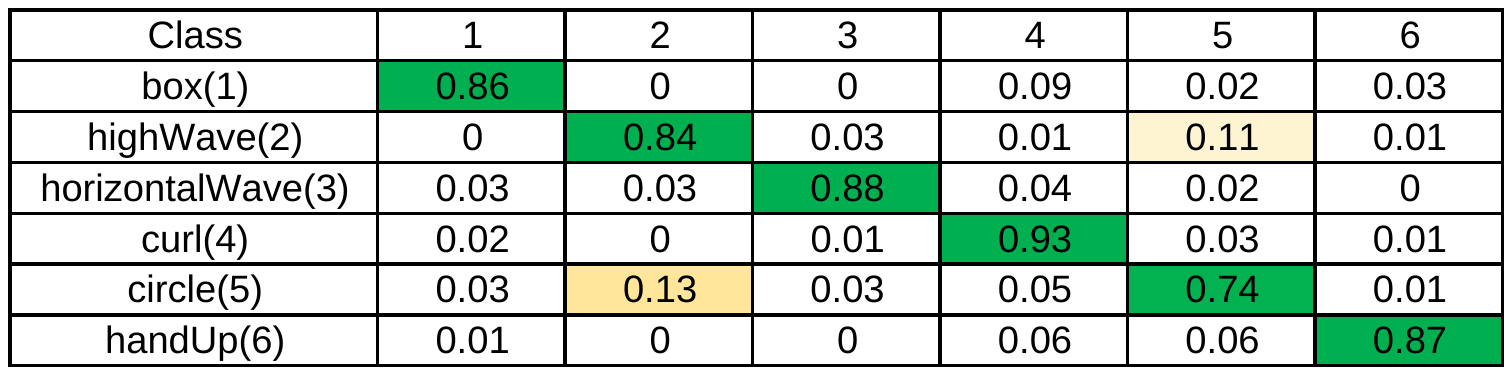}
\end{center}
\vspace{-1.0em}
\captionsetup{font={scriptsize}}
\caption{\textcolor[rgb]{0,0,0}{Confusion matrix of our method on the SmartHome dataset}}\label{491}\vspace{0em}
\end{figure}

\setlength{\tabcolsep}{2pt}
\begin{table}[t]
\renewcommand\arraystretch{0.9}
\centering
\scriptsize
\captionsetup{font={scriptsize}}
\caption{\textcolor[rgb]{0,0,0}{Comparison of recognition accuracy on SmartHome dataset between our method and previous approaches. ``background removal" refers to our method.}}\label{48}
\vspace{0em}
\footnotesize
\begin{tabular}{lcr}
\hline
         Methods                            &Accuracy  &   Details     \\
\hline
    \rowcolor[HTML]{FFCCC9}HON4D (without background removal) \cite{oreifej2013hon4d} &66.72\% &Oreifej \textsl{et al.} (2013) \\
    \rowcolor[HTML]{FFCCC9}HON4D (with background removal) \cite{oreifej2013hon4d} &78.23\% &Oreifej \textsl{et al.} (2013) \\
    \rowcolor[HTML]{FFCCC9}DSTIP+DCSF \cite{xia2013spatio} & 80.80\% &Xia \textsl{et al.} (2013) \\\rowcolor[HTML]{FFCCC9}{\color[HTML]{FE0000}\textbf{{M3DLSK+STV}}} &\textcolor[rgb]{1,0,0}{\textbf{85.31\%}} & Two scales\\
\hline
\end{tabular} \vspace{-0em}
\end{table}

\begin{figure}[t]
\begin{center}
\includegraphics[width=.85\linewidth]{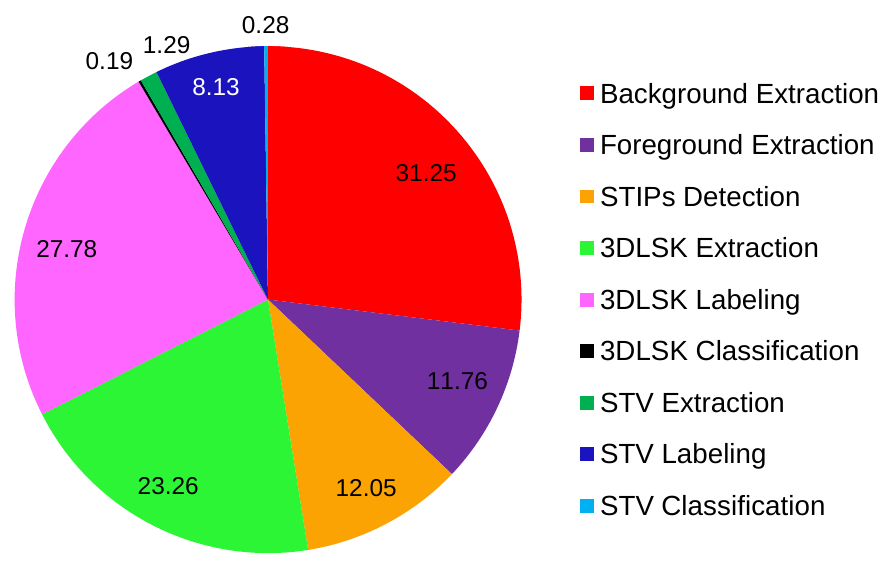}
\end{center}
\vspace{-1.0em}
\captionsetup{font={scriptsize}}
\caption{\textcolor[rgb]{0,0,0}{Evaluation of the time cost on the UTKinect-Action dataset}}\label{49}
\end{figure}

\setlength{\tabcolsep}{2pt}
\begin{table}[t]\footnotesize
\centering
\captionsetup{font={scriptsize}}
\vspace{-.5em}
\caption{\textcolor[rgb]{0,0,0}{Time cost comparison of background modeling on UTKinect-Action dataset}}\label{50}
\begin{tabular}{|c|c|}
\hline
         Background modeling method     &Time (ms/frame)     \\
\hline
Mean filter \cite{Benezeth2008Review} & 5.30 \\
\hline
ViBe \cite{Barnich2011ViBe} & 884.96 \\
\hline
{\color[HTML]{FE0000}\textbf{{3DLSK+STV}}} & 31.25 \\
\hline
\end{tabular}
\end{table}

\subsubsection{Robustness to partial occlusions}
\textcolor[rgb]{0,0,0}{To evaluate the robustness of the proposed framework to occlusion, we adopt the same settings as in \cite{wang2012robust} and divide the depth sequences from the MSRAction3D dataset into two parts in the x, y and t dimensions.
The whole sequences are divided into volumes, and eight kinds of occlusions are simulated by ignoring points that fall into specified volumes.
Fig. \ref{359} illustrates eight types of occlusions for the same depth sequence, where the shape of the actor is dramatically changed and some salient motion is also hidden.
The performance of our method is compared with that of the Random Occupancy Pattern (ROP) feature \cite{wang2012robust} in Table \ref{16}, where ``none" denotes the original MSRAction3D dataset.
Obviously, our method achieves higher precisions than ROP with all kinds of occlusions.
Note that sparse coding can improve the robustness of the given features to occlusions \cite{wang2012robust}.
Without applying sparse coding, our method outperforms ``ROP+sparse coding" under all types of occlusions.}

\subsubsection{Robustness to pepper noise}
\textcolor[rgb]{0,0,0}{Similar to \cite{Jetley20143D}, we simulate depth discontinuities in depth sequences by adding pepper noise in varying percentages (of the total number of image pixels) to depth images, as shown in Fig. \ref{355} (a). Despite the effects of pepper noise, our 3DLSK+STV achieves more than $90\%$ recognition accuracy on the MSRAction3D dataset, as shown in Fig. \ref{355} (b). This result proves the robustness of our method to depth discontinuities. The reason is that local noise are suppressed by our STIP detectors. Moreover, the 3DLSK descriptor suffers less from the effect of depth discontinuities. Motion History Template (MHT)+Binary Shape Templates (BST) \cite{Jetley20143D} is a global feature. Our method achieves higher accuracies than accuracies achieved in \cite{Jetley20143D}, which reflects the descriptive power of combining both local and global features.}

\setlength{\tabcolsep}{2pt}
\begin{table}[!htbp]\footnotesize
\centering
\captionsetup{font={scriptsize}}
\vspace{-.5em}
\caption{\textcolor[rgb]{0,0,0}{Testing speed comparison of different methods on the MSRDailyActivity3D dataset. The time cost of Depth Gradients+RDF is reported by \cite{rahmani2014real}.}}\label{51}
\begin{tabular}{|c|c|c|}
\hline
         Method     & Accuracy &Time (ms/frame)     \\
\hline
HON4D \cite{oreifej2013hon4d} & 80.00\% & 114.67\\
\hline
Depth Gradients+RDF \cite{rahmani2014real} &81.25\% &  8.93 \\
\hline
DSTIP+DCSF \cite{xia2013spatio} & 83.60\% & 96.53\\
\hline
{\color[HTML]{FE0000}\textbf{{M3DLSK+STV}}} & 91.00\% & 74.64 \\
\hline
\end{tabular}
\end{table}

\subsubsection{Smart home control}
The SmartHome dataset \footnote{Dataset provided in \url{http://pan.baidu.com/s/1gf9ZmCz}} was designed by our lab for smart home control and contains six gestures: ``box", ``highWave", ``horizontalWave", ``curl", ``circle", and ``handUp".
Each action is performed six times (three times for each hand) by nine subjects in five situations: ``sit", ``stand", ``with a pillow", ``with a laptop", and ``with a person", resulting in 1620 depth sequences.
\textcolor[rgb]{0,0,0}{Several action frames from the SmartHome dataset are shown in Fig. \ref{46}, where inter-class similarities among different actions and intra-class dissimilarities among same type of actions are observed.
What is more, different poses and distractors (e.g., ``a pillow" and ``a computer"), shown in Fig. \ref{47}, further increase the intra-class dissimilarities.} We use cross-subject validation with subjects 1, 2, 3, and 4 for training and subjects 5, 6, 7, 8, and 9 for testing.

M3DLSK+STV achieves an accuracy of $85.31\%$ on the SmartHome dataset, where the M3DLSK descriptor is implemented with two scales and all parameters are selected to achieve optimal performance.
Fig. \ref{491} shows the confusion matrix of our method, where actions ``highWave" and ``circle" have maximum confusion with each other since both actions share the motion ``raise a hand and then drop it down".
\textcolor[rgb]{0,0,0}{In Table \ref{48}, our method is compared with two popular local-feature-based methods: Histogram of Oriented 4D Normals (HON4D) \footnote{Code provided in \url{http://www.cs.ucf.edu/~oreifej/HON4D.html}} and STIPs from depth videos (DSTIP)+depth cuboid similarity feature (DCSF) \footnote{Code provided in \url{http://cvrc.ece.utexas.edu/lu/index.html}}.
Our method achieves performance that is $7.08\%$ higher than \cite{oreifej2013hon4d} and $4.51\%$ higher than \cite{xia2013spatio}, which verifies the merit of combining both local and global features for solving inter-class similarities.}

\subsection{Computational Cost}
We evaluate the time cost of our method on a 2.5 GHz machine with 8 GB RAM using Matlab R2012a.
Default parameters are used for implementing the 3DLSK and STV descriptors.
Fig. \ref{49} reports the average computational time costs associated with all steps of our method, e.g., foreground extraction, the 3DLSK descriptor and the STV descriptor, on the UTKinect-Action dataset.
The background extraction incurs the maximum time cost: 31.25 milliseconds per frame.
\textcolor[rgb]{0,0,0}{In Table \ref{50}, the background modeling method is compared with Mean filter \cite{Benezeth2008Review} and ViBe \cite{Barnich2011ViBe}. We claim that our method achieves better performance than Mean filter and ViBe, while incurring smaller time cost. The labeling step and the feature extraction step of the 3DLSK descriptor also incur large time costs: 27.78 milliseconds per frame and 23.26 milliseconds per frame, respectively.
In Table \ref{51}, we compare the testing speed of our method with those of HON4D \cite{oreifej2013hon4d} and Depth Gradients+RDF \cite{rahmani2014real}. We test all testing sequences and report the average time cost for each frame.
Our method achieves higher accuracy than that in \cite{rahmani2014real} but incurs larger time cost.}

\section{Conclusions and future work}
In this paper, we propose to use motion-based and shape-based STIPs to comprehensively characterize the depth action cues for effective action recognition. To tune this action representation for our problem, we propose two new feature descriptors M3DLSK and STV to jointly model the local motion structure and global shape distribution using motion-based and shape-based STIPs. A two-layer BoVW model is finally presented to represent 3D actions by fusing local appearances and global distributions of STIPs.
Additionally, we propose a new background modeling method as a preprocessing step, which facilitates the detection of motion-based and shape-based STIPs.
Experimental results on the MSRAction3D and MSRGesture3D datasets show that our method outperforms existing depth-based methods on benchmark datasets that are designed for either action recognition or gesture recognition.
Our method shows robustness to viewpoint changes to some extent, which is verified on the UTKinect-Action dataset.
Our method can efficiently encode human-object interactions and actions with slight motions, which is verified on the MSRDailyActivity3D dataset.
Our method also shows robustness to partial occlusions and depth discontinuities; this is verified on two modified MSRAction3D datasets.
Results on SmartHome dataset show the robustness of our method to intra-class dissimilarities and inter-class similarities.
Future work will focus on developing a real-time 3D action recognition system for daily assisted living based on current work.

\bibliography{refs}{}
\bibliographystyle{ieeetr}

\vspace{-4em}
\begin{IEEEbiography}[{\includegraphics[width=.78in,height=1in]{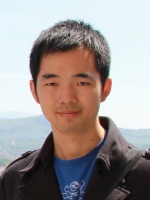}\vspace{2.5em}}]{Mengyuan Liu}
received the B.E. degree in intelligence science and technology from Nankai University (NKU), China, in 2012, and received the Doctor degree in the School of EE\&CS, Peking University (PKU), China, in 2017. He is a Post-Doc in the School of Electrical and Electronic Engineering, Nanyang Technological University (NTU), Singapore. His research interests include human action recognition and detection.
\end{IEEEbiography}

\vspace{-5.5em}
\begin{IEEEbiography}[{\includegraphics[width=.78in,height=1.2in]{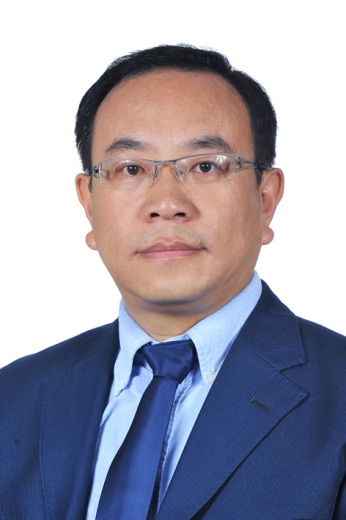}}]{Hong Liu}
received the Ph.D. degree in mechanical electronics and automation in 1996, and serves as a Full Professor in the School of EE\&CS, Peking University (PKU), China. Prof. Liu has been selected as Chinese Innovation Leading Talent supported by “National High-level Talents Special Support Plan” since 2013.
He is also the Director of Open Lab on Human Robot Interaction, PKU, his research fields include computer vision and robotics, image processing, and pattern recognition. Dr. Liu has published more than 150 papers and gained Chinese National Aero-space Award, Wu Wenjun Award on Artificial Intelligence, Excellence Teaching Award, and Candidates of Top Ten Outstanding Professors in PKU. He is an IEEE member, vice president of Chinese Association for Artificial Intelligent (CAAI), and vice chair of Intelligent Robotics Society of CAAI. He has served as keynote speakers, co-chairs, session chairs, or PC members of many important international conferences, such as IEEE/RSJ IROS, IEEE ROBIO, IEEE SMC and IIHMSP, recently also serves as reviewers for many international journals such as Pattern Recognition, IEEE Trans. on Signal Processing, and IEEE Trans. on PAMI.
\end{IEEEbiography}

\vspace{-3.5em}
\begin{IEEEbiography}[{\includegraphics[width=.78in,height=0.9in]{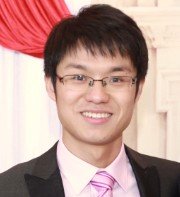}}]{Chen Chen}
received the B.E. degree in automation from Beijing Forestry University, Beijing, China, in 2009 and the MS degree in electrical engineering from Mississippi State University, Starkville, in 2012 and the PhD degree in the Department of Electrical Engineering at the University of Texas at Dallas, Richardson, TX in 2016. He is a Post-Doc in the Center for Research in Computer Vision at University of Central Florida (UCF). His research interests include signal and image processing, pattern recognition and computer vision.
\end{IEEEbiography}
\end{document}